\documentclass{article}

\usepackage{arxiv}

\usepackage{arxiv}
\usepackage{microtype}
\usepackage{graphicx}
\usepackage{caption}
\usepackage{subcaption}
\usepackage{booktabs} 
\usepackage{hyperref}

\usepackage{amsmath}
\usepackage{amssymb}
\usepackage{amsthm}
\usepackage{mathabx}
\usepackage{bbm}
\usepackage{float}
\pdfoutput=1
\usepackage{soul}
\PassOptionsToPackage{margin=1in}{geometry}

\usepackage{color}
\definecolor{blue}{rgb}{0,0,1}

\usepackage[capitalize,noabbrev]{cleveref}
\usepackage[textsize=tiny]{todonotes}
\DeclareMathOperator*{\argmin}{argmin}

\newcommand{\ep}{\mathbb{E}}

\newcommand\expe[1]{\mathop{\ep}_{#1}}
\newcommand\tin[1]{\text{\textit{\tiny{#1}}}}

\newcommand\bhf{\bar{\hat{f}}}
\newcommand\hfs{\hat{f_{\mathcal{S}}}}

\newcommand\hYsm{\hat{Y}_{\mathcal{S}_m}}

\newcommand\bhY{\bar{\hat{Y}}}
\newcommand\bhy{\bar{\hat{y}}}

\newcommand\Yhs{\hat{Y}_{\mathcal{S}}}
\newcommand\bx{\mathbf{x}}
\newcommand\cX{\mathcal{X}}

\DeclareMathAlphabet\mathbfcal{OMS}{cmsy}{b}{n}

\newcommand\bhYmm{\bhY_{\frac{m_1}{m_0}}}
\newcommand\hYmm{\hat{Y}_{\mathcal{S}_{\frac{m_1}{m_0}}}}
\newcommand\bhYmmp{\bhY_{\frac{m^p_1}{m^p_0}}}

\theoremstyle{plain}
\newtheorem{theorem}{Theorem}[section]

\theoremstyle{definition}
\newtheorem{definition}[theorem]{Definition}

\theoremstyle{remark}

\title{On the Origins of Sampling Bias: Implications on Fairness Measurement and Mitigation}

\author{
  Sami Zhioua\\
  University of Doha for Science and Technology (UDST) \\
  Doha, Qatar\\
  \texttt{sami.zhioua@udst.edu.qa} \\
   \And
  Ruta Binkyte \\
  CISPA Helmholtz Center for Information Security \\
  Saarbrücken, Germany\\
  \texttt{ruta.binkyte-sadauskiene@cispa.de} \\
   \And
  Ayoub Ouni \\
  Université Claude Bernard Lyon 1 \\
  Lyon, France\\
  \texttt{ayoub.ouni@etu.univ-lyon1.fr} \\
   \And
  Farah Barika Ktata \\
  MIRACL, ISSAT, Université de Sousse \\
  Sousse, Tunisia\\
  \texttt{farah.barika.ktata@issatso.u-sousse.tn} \\
}

\begin{document}
\maketitle

\begin{abstract}

Accurately measuring discrimination is crucial to faithfully assessing fairness of trained machine learning (ML) models. Any bias in measuring discrimination leads to either amplification or underestimation of the existing disparity.  
Several sources of bias exist and it is assumed that bias resulting from machine learning is born equally by different groups (e.g. females vs males, whites vs blacks, etc.). If, however, bias is born differently by different groups, it may exacerbate discrimination against specific sub-populations. 
Sampling bias, in particular, is inconsistently used in the literature to describe bias due to the sampling procedure. In this paper, we attempt to disambiguate this term by introducing clearly defined variants of sampling bias, namely, sample size bias ({\em SSB}) and underrepresentation bias ({\em URB}). Through an extensive set of experiments on benchmark datasets and using mainstream learning algorithms, we expose relevant observations in several model training scenarios. The observations are finally framed as actionable recommendations for practitioners.
\end{abstract}


\section{Introduction}
\label{sec:intro}

With the growing reliance on machine learning (ML) systems for critical decisions (e.g., hiring, college admissions, and security screening), ensuring fairness is paramount. Unfair outcomes can amplify discrimination against individuals or sub-populations, creating a vicious cycle with severe consequences. 
One of the most known examples of ML discrimination is COMPAS software~\cite{COMPAS} used by several states in the US to help predict whether a defendant will recidivate in the next two years if she is released. The second example is related to face recognition technology (FRT). Buolamwini et al.~\cite{buolamwini2018gender} found that several commercial FRT software have a significantly lower accuracy for individuals belonging to a specific sub-population, namely, dark-skinned females. 

Discrimination in ML arises from two main sources: biased labels due to societal inequalities, and algorithmic disparities in assigning labels accurately across sensitive groups. The latter most often occurs due to sampling bias, namely,  when ML models are trained on limited or imbalanced datasets. This produces inaccurate models and the inaccuracy will typically be born differently by different sub-populations which leads to discrimination. This paper formalizes two key forms of sampling bias: sample size bias (SSB), where training data has limited samples proportionally representing the population, and underrepresentation bias (URB), where certain sub-populations are disproportionately represented. While these biases are commonly referred to under various terms in the literature (e.g., representation bias, data imbalance bias), they lack consistent definitions, which this work aims to address.


Our work builds on prior research on auditing group-wise performance disparities in ML models~\cite{vogel2021learning,saleiro2018aequitas,fletcher2021addressing,farrand2020neither,chen2018,yan2020fair}. Recent advancements~\cite{cherian2024statistical,rauba2024context} emphasize the need to assess model performance across subpopulations while addressing challenges such as multiple testing and distribution shifts. These studies highlight the importance of ensuring that discrimination observed in test datasets generalizes to deployment populations.

First, we build on foundational work~\cite{dietterich1995,domingos2000,chen2018} to formally define {\em SSB} and {\em URB}, providing a framework to isolate and analyze their effects. Then, through empirical analysis on benchmark datasets, we examine how these biases distort measurements of discrimination. We conduct extensive experiments using three widely used benchmark datasets, six fairness metrics, and five commonly applied classifiers for tabular data. To evaluate discrimination we use fairness metrics consistent with the second source of algorithmic discrimination, namely disparities in assigning labels for sensitive groups. This type of discrimination is best captured by the metrics that take into account the correspondence between the true and assigned labels, such as Equal Opportunity (EO), False Positive Rate (FPR), and others. Our goal is to evaluate, how robust the metrics are to {\em SSB} and {\em URB}. More precisely, how informative is the evaluation of discrimination on the biased sample to the expected discrimination in the population after the model is deployed? 

Next, we explore the impact of sampling bias on the bias mitigation techniques. Our study explores the effectiveness of pre-processing and in-processing~\cite{dunkelau2019fairness,hort2024bias,siddique2023survey} in the presence of sampling bias. Another common approach for mitigating sampling bias, particularly related to computer vision, is data augmentation(~\cite{pastaltzidis2022data,yucer2020exploring, zhang2020towards,xu2020investigating,Wang2018BalancedDA,wang2020towards}). We perform experiments using random and selective data augmentation techniques and analyze their effect on discrimination in the presence of varying levels of {\em SSB}, {\em URB}, and data disparity. 

Our findings provide insights into the metrics and mitigation techniques that perform best in the presence of {\em SSB} or {\em URB}. First, when data is biased (high discrimination) and a model is trained using a small or imbalanced dataset, fairness metrics tend to underestimate the true discrimination value rather than exacerbate it. Second, fairness metrics which combine sensitivity and specificity ({\em AUC}, {\em ZOL}, etc.) are less sensitive to {\em SSB} and {\em URB} and hence provide a more reliable assessment of discrimination. 

For pre/in-processing approaches, the main finding is that their efficiency in mitigating bias is diluted by small ({\em SSB}) or imbalanced ({\em URB}) training datasets. For data augmentation, our experiments show that augmenting data by collecting more samples from underrepresented groups can amplify discrimination when the new samples reflect the same biased distribution. Instead, balancing data with respect to outcomes proves more effective. Finally, all findings of this work are compiled and framed as high-level and actionable recommendations for practitioners.

The key findings of this paper are the following:
\begin{itemize}
\item Discrimination defined in terms of cost/accuracy metrics that consider a trade-off between precision and recall (e.g. $AUC$ and $ZOL$) are more resilient to limited size or imbalanced training sets.
\item The effect of the SSB and URB on the measurement of discrimination is directly dependent on the initial disparity in the data.
\item In presence of underrepresented groups, collecting more data samples for the underrepresented group typically amplifies discrimination rather than reduces it.
\end{itemize}









\section{Preliminaries}

Let $\mathcal A$ be a supervised learning algorithm for learning an unknown function $f: \mathbfcal{X} \mapsto \mathcal{Y}$ where $\mathbfcal X$ is the input variables space and $\mathcal Y$ is the outcome space. Without loss of generality, the outcome random variable $Y$ is assumed to be binary ($\mathcal{Y} = \{0,1\}$, e.g. accepted/rejected). 
Let $\mathcal{S} = \{(\mathbf{x}_i,y_i=f(\mathbf{x}_i))\}, i=1\ldots m$, be a training sample of size $m$. 
Based on the data sample $\mathcal S$, algorithm $\mathcal A$ learns a function $\mathcal{A}(\mathcal{S}) = \hat{f}_{\mathcal{S}}^{\mathcal{A}}$. Let $\hat{Y}_{\mathcal{S}}^{\mathcal{A}}$ be the predicted outcome random variable such that $\hat{f}_{\mathcal{S}}^{\mathcal{A}}(\mathbf{x}_i) = \hat{y_i}$. When there is no ambiguity, we refer to $\hat{Y}_{\mathcal{S}}^{\mathcal{A}}$ and $\hat{f}_{\mathcal{S}}^{\mathcal{A}}$ simply as $\hat{Y}$ (or $\Yhs$) and $\hat{f}$ (or $\hat{f}_{\mathcal{S}}$).

Given the true value $y$ and the prediction $\hat{y}$, $L(y,\hat{y})$ represents the loss incurred by predicting $\hat{y}$ while the true outcome is $y$. A commonly used loss function for regression problems is the squared loss defined as $L^{SL}(\hat{y},y) = (\hat{y}-y)^2$. Other loss functions that will be considered in this paper are the absolute loss $L^{AL}(\hat{y},y) = |\hat{y}-y|$ and the zero-one loss $L^{ZO}(\hat{y},y) = 0$ if $\hat{y} = y$, and $1$ otherwise.

Based on a loss function, we define two special predictions, namely, the main prediction for a learning algorithm $\mathcal{A}$ and the optimal prediction. 

Given a learning algorithm $\mathcal{A}$ and a set of training samples $\mathfrak{S} = \{\mathcal{S}_1, \mathcal{S}_2, \ldots\}$, the main prediction random variable $\bhY^{\mathcal{A}}_{\mathfrak{S}}$ ($\bhy = \bhf_{\mathfrak{S}}^\mathcal{A}(\bx)$) represents the prediction that minimizes the loss across all training sets in $\mathfrak{S}$. That is, 
$$\bhf^{\mathfrak{S}}_{\mathcal{A}}(\bx) = \argmin_{f'} \ep_{\mathcal{S} \in \mathfrak{S}} [L(\hfs(\bx),f'(\bx)].$$ 
When there is no ambiguity, we refer to $\bhY^{\mathcal{A}}_{\mathfrak{S}}$ and $\bhf_{\mathfrak{S}}^{\mathcal{A}}(\bx)$ simply as $\bhY$ and $\bhf(\bx)$.
Typically, the main prediction corresponds to the average prediction 
across all training sets in $\mathfrak{S}$. That is,
\begin{equation}\bhf(\bx) =  \displaystyle \expe{\mathcal{S} \in \mathfrak{S}} \hat{f}_{\mathcal{S}}(\mathbf{x})\footnote{We exceptionally use the expectation on a function, instead of a random variable.}.\end{equation}

The optimal prediction $Y^*$ ($y^* = f^{*}(\bx)$) is the prediction that minimizes the loss across all possible predictors. That is,
$$f^{*}(\bx) = \argmin_{f'} \ep[L(f(\bx),f'(\bx)].$$
It is important to note that $f^{*}$ is independent of the learning algorithm $\mathcal{A}$.

Assume that the sensitive attribute $A$ is a binary variable with possible values $A=a_0$ and $A=a_1$, each representing a different group (e.g. male vs female, black vs white, etc.). Let $G_0$ and $G_1$ denote these groups. That is, $G_0 = \{\bx \in \cX | A=a_0\}$ and $G_1 = \{\bx \in \cX | A=a_1\}$. Discrimination between $G_0$ and $G_1$ can be defined in terms of the disparity in prediction accuracy. Let $C_a^{\bullet}(\hat{Y})$ denote the accuracy/cost of prediction $\hat{Y}$ for group $A=a$. For classification problems, we consider four metrics, namely, false positive rate ($FPR$), false negative rate ($FNR$), true positive rate ($TPR$), and zero one loss ($ZOL$). 
These metrics are defined as follows: 
\begin{itemize}
    \item[$\circ$] $C^{\tin{FPR}}_a(\hat{Y}) = \ep[\hat{Y} | Y=0, A=a]$
    \item[$\circ$] $C^{\tin{FNR}}_a(\hat{Y}) = \ep[1-\hat{Y} | Y=1, A=a]$
    \item[$\circ$] $C^{\tin{TPR}}_a(\hat{Y}) = \ep[\hat{Y} | Y=1, A=a]$
    \item[$\circ$] $C^{\tin{ZOL}}_a(\hat{Y}) = \ep[\mathbbm{1}[\hat{Y}\neq Y]| A=a]$
\end{itemize}

Discrimination $Disc^{\bullet}$ can be defined as the difference in $C_a^{\bullet}$ between the two sensitive groups. For instance $Disc^{\tin{FPR}}(\hat{Y}) = C_{a_1}^{\tin{FPR}}(\hat{Y}) - C_{a_0}^{\tin{FPR}}(\hat{Y})$. Notice that $Disc^{\tin{TPR}}(\hat{Y})$ corresponds to discrimination according to equal opportunity~\cite{hardt2016equality} and that $Disc^{\tin{TPR}}(\hat{Y}) = - Disc^{\tin{FNR}}(\hat{Y})$ as $TPR = 1 - FNR$. In the rest of the paper, we use $Disc^{\tin{TPR}}(\hat{Y})$ and $Disc^{\tin{EO}}(\hat{Y})$ interchangeably. In addition, for reference, we use $Disc^{\tin{SD}}(\hat{Y}) = \ep[\hat{Y}|A=a_1] - \ep[\hat{Y}|A=a_0]$ to denote statistical disparity~\cite{dwork2012fairness}.


\section{Sample Size and Underrepresentation Biases}


Typically, the size of the data used to train an ML model has a significant impact on the accuracy of the obtained model. However, it is generally assumed that the loss in accuracy is equally born by the different segments of the data. As it is not usually the case, we define sample size bias (SSB) as the bias resulting from training a model with a given data size.


Let $\mathfrak{S_m} = \{\mathcal{S}_1, \mathcal{S}_2, \ldots\}$ be the set of samples of size $m$, and let $\hat{f}_{\mathcal{S}_1}, \hat{f}_{\mathcal{S}_2}, \ldots$ be the models produced by applying the learning algorithm $\mathcal A$ on each sample ($\mathcal{A}(\mathcal{S}_1)=\hat{f}_{\mathcal{S}_1}$, etc.). Let $\bhY^{\mathcal{A}}_{\mathfrak{S_m}}$ ($\bhy_m = \bhf_{\mathfrak{S}_m}^{\mathcal{A}}(\bx)$) be the main prediction obtained using the set of training sets $\mathfrak{S_m}$. That is,
\begin{equation}\bhf_{\mathfrak{S}_m}^{\mathcal{A}}(\bx) = \argmin_{f'} \ep_{\mathcal{S} \in \mathfrak{S}_m} [L(\hfs(\bx),f'(\bx)].
\end{equation}
When there is no ambiguity, we refer to $\bhY^{\mathcal{A}}_{\mathfrak{S_m}}$ and $\bhf_{\mathfrak{S}_m}^{\mathcal{A}}$ simply as $\bhY_m$ and $\bhf_m$.

\begin{definition}
Given a positive number $m > 0$ representing the training set size, sample size bias is the difference in discrimination due to the training set size:
\begin{equation}
    SSB^{\bullet}(\mathcal{A},m) = Disc^{\bullet}(\bhY_m) - Disc^{\bullet}(\bhY_{\infty}) \label{eq:ssb0}
\end{equation}
where $Disc^{\bullet}(\bhY_{\infty}) = \displaystyle \lim_{m\to \infty} Disc^{\bullet}(\bhY_m) $ and $\bullet$ is a placeholder for the accuracy/cost metric ($FPR, FNR, EO, ZOL,$ or $MSE$ for regression problems). As a metric that combines both specificity ($FPR$) and sensitivity ($TPR$), we use also $AUC$ (area under the curve)\footnote{Other metrics combining specificity and sensitivity include $F_1$ score and balanced accuracy ($BA$)}. 
\end{definition}

As $SSB$ is defined in terms of an infinite size training set ($\bhY_{\infty}$), we consider an alternative definition in terms of $M$, the size of the largest training set available:
\begin{equation}
    SSB_M^{\bullet}(\mathcal{A},m) = Disc^{\bullet}(\bhY_m) - Disc^{\bullet}(\bhY_M) \label{eq:ssb1}
\end{equation}

Another variant of $SSB$ can be defined based on a specific training set $\mathcal{S}_m$ of size $m$ as follows:
\begin{equation}
    SSB_M^{\bullet}(\mathcal{A},\mathcal{S}_m) = Disc^{\bullet}(\hYsm) - Disc^{\bullet}(\bhY_M) \label{eq:ssb2}
\end{equation}



When sampling a training set from a population, it is generally assumed that the generated sample is balanced. Data is balanced if all classes are proportionally represented and is imbalanced if it suffers from severe class distribution skews~\cite{he2009learning}. For instance, if one class label is overrepresented at the expense of another underrepresented class label. If data is imbalanced in the sensitive groups (e.g. male vs female, blacks vs whites, etc.), it can have significant impact on the disparity of accuracies and consequently on discrimination between sensitive groups. We define underrepresentation bias (URB) as the bias resulting from a disparity in representation between the sensitive groups. 

Let $\mathfrak{S^{\frac{m_1}{m_0}}_m}$ be the set of samples of size $m$ with $m_0$ and $m_1$ items from $G_0$ and $G_1$ respectively. That is, for $\mathcal{S} \in \mathfrak{S^{\frac{m_1}{m_0}}_m}$, $|\{\bx \in \mathcal{S} | A=a_0\}| = m_0$, $|\{\bx \in \mathcal{S} | A=a_1\}| = m_1$, and $m_0 + m_1 = m = |\mathcal{S}|$. We use the simpler notation $\bhY_{\frac{m_1}{m_0}}$ to refer to $\bhY_{\mathfrak{S^{\frac{m_1}{m_0}}_m}}^{\mathcal{A}}$.

\begin{definition}
\label{def:urb}
Given, $m, m_0, m_1 > 0$ such that $m_0 + m_1 = m$, underrepresentation bias is the difference in discrimination due to the disparity in sample sizes compared to the population ratio:

\begin{equation}
 URB^{\bullet}(\mathcal{A},m_0,m_1) = Disc^{\bullet}(\bhYmm) - Disc^{\bullet}(\bhYmmp)   
\end{equation}
where $Disc^{\bullet}(\bhY_{{m^p_1}/{m^p_0}})$ is the discrimination of the prediction based on a model trained using only samples from $\mathfrak{S^{{m^p_1}/{m^p_0}}_m}$, and the ratio $\frac{m^p_1}{m^p_0}$ is the same as the ratio in the population ($\frac{m^p_1}{m^p_0} \approx \frac{|G_1|}{|G_0|}$).
\end{definition}

Similar to $SSB_M^{\bullet}(\mathcal{A},\mathcal{S}_m)$ (Equation~\ref{eq:ssb2}), a variant of $URB$ can be defined based on a specific training set $\mathcal{S}_{\frac{m_1}{m_0}} \in \mathfrak{S^{\frac{m_1}{m_0}}_m}$ as follows:
\begin{equation}
    URB^{\bullet}(\mathcal{A},\mathcal{S}_{\frac{m_1}{m_0}}) = Disc^{\bullet}(\hYmm) - Disc^{\bullet}(\bhYmmp) \label{eq:urb2}
\end{equation}

\section{Experimental Setup}
\label{sec:exp}

\begin{figure}
    \hspace*{-1cm}
    \includegraphics[height=1.6in, width=8.2in]{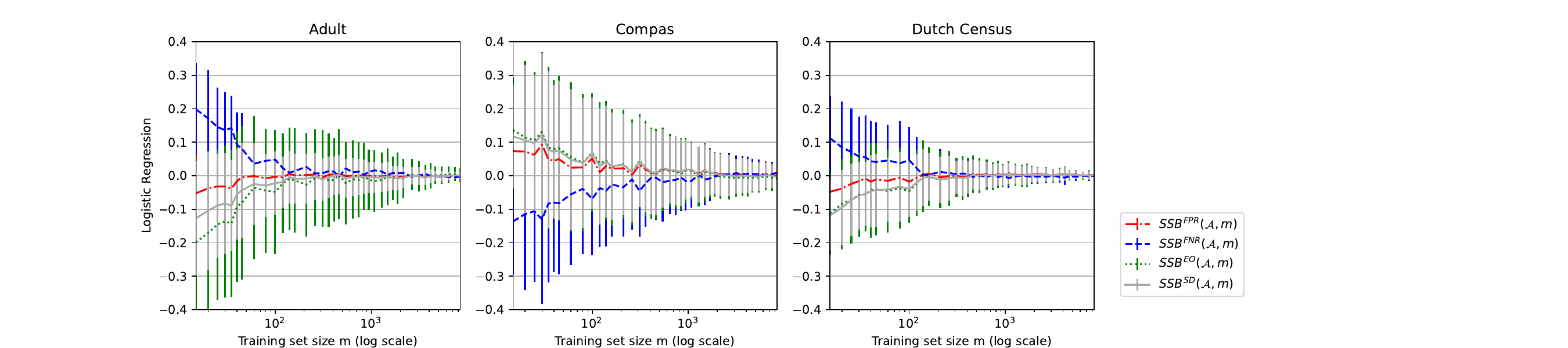}
    \hspace*{-1cm}
    \includegraphics[height=1.6in, width=8.2in]{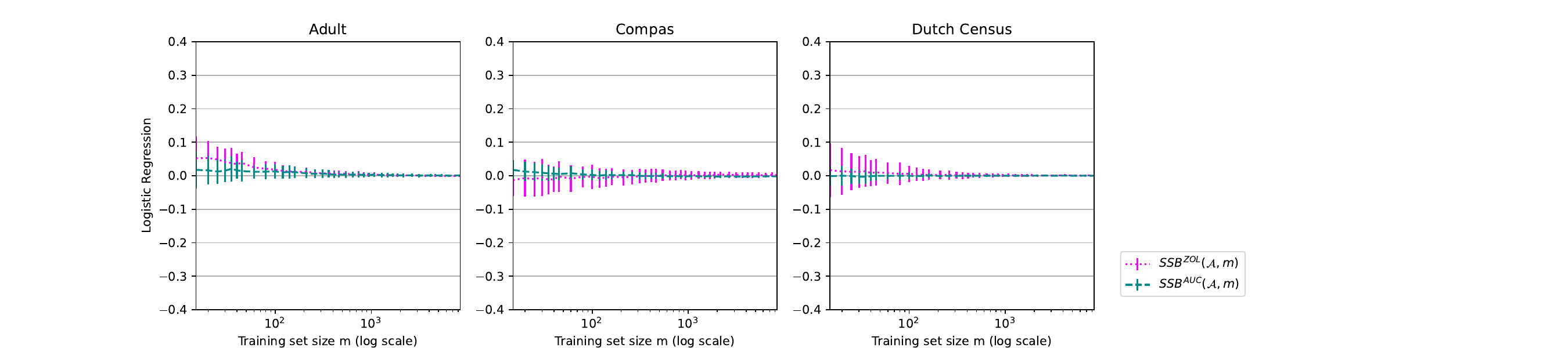}
    \caption{Magnitude of sample size bias (SSB) for increasing size of the training data.}
    \label{fig:ssb_lg_orig}
\end{figure}

\label{app:methods}

This study investigates the impact of sample size bias ({\em SSB}) and underrepresentation bias ({\em URB}) on fairness evaluations, particularly in datasets with varying levels of imbalance. {\em SSB} is analyzed by training ML models with datasets of increasing sizes, while {\em URB} is studied by varying the representation of sensitive groups in training data. Model performance is evaluated on population-sized test sets to demonstrate how these biases distort fairness assessments when fairness metrics are applied to smaller test sets or test sets with imbalances different from that of the population. The study also examines the effect of bias mitigation techniques (pre-processing and in-processing) and data augmentation on discrimination. 


\subsection{Sample Size Bias (SSB)}

\label{sec:exp_ssb}
To measure SSB, ML models are trained on datasets with increasing sample sizes (e.g., from 10 to 2000 samples for COMPAS). For each size $m$, 30 samples are drawn, and a separate model is trained on each. 
To analyze further the behavior of {\em SSB}, we focus on the case of extreme imbalance between the protected groups. For example, for the adult dataset, we consider a version in which the positive outcome rate is $0.9$ for the privileged group (males) and $0.1$ for the unprivileged group (females).

\subsection{Uderrepresentation Bias (URB)}


The aim for underrepresentation bias experiment is to observe the magnitude of {\em URB} while the ratio of the sensitive groups in the training set is changing. We consider different values of the splitting $\frac{m_1}{m_0}$ (see Definition~\ref{def:urb}) (e.g. $0.1$ vs $0.9$, $0.2$ vs $0.8$, etc.). However, as {\em URB} is more significant for extreme disparities in group size, we focus more on extreme splitting values (e.g. $0.001$ vs $0.99$, $0.002$ vs $0.98$, etc.). A similar behavior has been observed previously by Farrand et al.~\cite{farrand2020neither}. Assuming a fixed sample size (e.g. $1000$), for each splitting value, we sample the data so that the proportions of sensitive groups (e.g. male vs female) match the splitting value. Similarly to the {\em SSB} experiment, we repeat the sampling several times ($30$ by default) for the same splitting value. Then, we train a different model using each one of the samples so that we obtain $30$ models for each splitting value $\frac{m_1}{m_0}$. The discriminations obtained using the different models are then averaged across all models.
Similarly to {\em SSB} experiment, to analyze further the behavior of {\em URB} in presence of significant discrimination, we focus on the case of extreme imbalance between the protected groups. For example, for the adult dataset, we consider a version in which the positive outcome rate is $0.9$ for the privileged group (males) and $0.1$ for the unprivileged group (females).



\subsection{Bias Mitigation}
Next, we focus on the effect of both types of bias ({\em SSB} and {\em URB}) on discrimination when bias mitigation approaches are used in model training. Several bias mitigation approaches have been developed in the few last years~\cite{dunkelau2019fairness,hort2024bias,siddique2023survey}. Most of these approaches fall into two categories. First pre-processing techniques which try to transform the data so that the underlying discrimination is removed. Second, in-processing techniques which try to modify the learning algorithms in order to remove discrimination during the model training process. We consider one representative technique from each approach. First, Reweighing~\cite{kamiran2012data} tries to mitigate the effects of discrimination on the learned model by adjusting the weights of training samples based on group representation. This method ensures that groups that are underrepresented in the training data are given a higher weight during the learning process, thus improving the model's ability to learn from these groups effectively. We applied reweighing during the preprocessing stage to all datasets before model training. Second, GerryFairClassifier~\cite{kearns2019empirical} which introduces fairness constraints directly into the learning process. This in-processing method works by adjusting the optimization objective of the model to account for fairness criteria, effectively reducing disparities across sensitive groups while still striving for predictive accuracy.


\subsection{Data Augmentation}

The natural approach to address sampling bias is to use more data for training, in particular for the under-represented groups. Obtaining more data is possible either through data augmentation or data collection. Data augmentation is the process of using the available data to generate more samples. In turn, this can be done in two ways: oversampling or creating synthetic samples. Oversampling consists in duplicating existing samples to balance the data. A simple variant is to randomly duplicate samples from the under represented group. Creating synthetic samples, on the other hand, is typically done using SMOTE~\cite{chawla2002smote}. 
Both techniques of data augmentation try to balance data by adding artificially generated samples. While this artificial manipulation may reduce discrimination between sensitive groups, it can lead to models which are not faithful to reality. When it is possible, collecting more data is more natural and reflects better reality. The approach is simple: if a sensitive group is under represented, collect more samples of that group. Unlike data augmentation, whose effect on discrimination has been the topic of a number of papers, in particular related to computer vision (e.g.~\cite{pastaltzidis2022data,yucer2020exploring, zhang2020towards,xu2020investigating,Wang2018BalancedDA,wang2020towards}), the impact of  of collecting more samples on discrimination has not been well studied in the literature. 

We devise simple experiments to observe the effect of populating the data with more samples collected from the same population as the existing data. Using the same benchmark datasets, the aim is to train models based on an increasing number of one protected group while keeping the other group portion unchanged. 
For instance, for the Dutch Census dataset, models are trained with $100$ privileged group samples and increasing protected group samples (from $2$ to $100$). We use 3-fold cross-validation and randomly generate $50$ different samples for every size value. In the selective data augmentation variant of the same experiment, additional samples are collected exclusively from the protected group with positive outcomes. This behavior is compared to two other augmentation strategies: randomly collecting more protected group samples and randomly collecting more unprotected group samples. In all three cases, the importance of the sensitive feature (e.g., sex) in prediction, as measured by SHAP explanations~\cite{lundberg2017unified}.

\section{Experimental Analysis}
In this section we provide the results of the experiments on {\em SSB}, { \em URB}, fairness pre-processing and in-processing algorithms, and data augmentation techniques.

\subsection{Sample Size Bias ({\em SSB})}
\label{sec:ssbexp}

\paragraph{\bf \textit{[Obs1] Magnitude of Sample Size Bias ({\em SSB})} \label{Obs1} }Figure~\ref{fig:ssb_lg_orig} shows the magnitude of SSB according to each metric, for each benchmark dataset, and using logistic regression. Notice that $SSB^{EO}$ and $SSB^{FNR}$ are symmetric because, as mentioned above, $FNR = 1 - TPR$ and hence $SSB^{EO} = - SSB^{FNR}$. Most of the plots exhibit an expected behavior of SSB. That is, the bias is significant when the models are trained using a limited size training set. The bias disappears gradually as the training set size increases. {{\em SSB}} behaves the same way for the other classifiers (Figures~\ref{fig:all_ssb1} and~\ref{fig:all_ssb2} in Appendix~\ref{sec:a_ssb}). More importantly, {\em SSB} results show that metrics that combine specificity and sensitivity ({\em AUC} and {\em ZOL}) are less sensitive to the training set size than the remaining metrics ($FPR$ and $EO$). A possible explanation is that for small training sets, it is more likely that a majority of the samples have the same outcome (positive or negative) which can boost precision on the expense of recall or the opposite. {\em AUC} and {\em ZOL} are not subject to such skewness since they consider the trade-off between precision and recall.

\begin{figure}
\small
    \centering
    \includegraphics[scale=0.34]{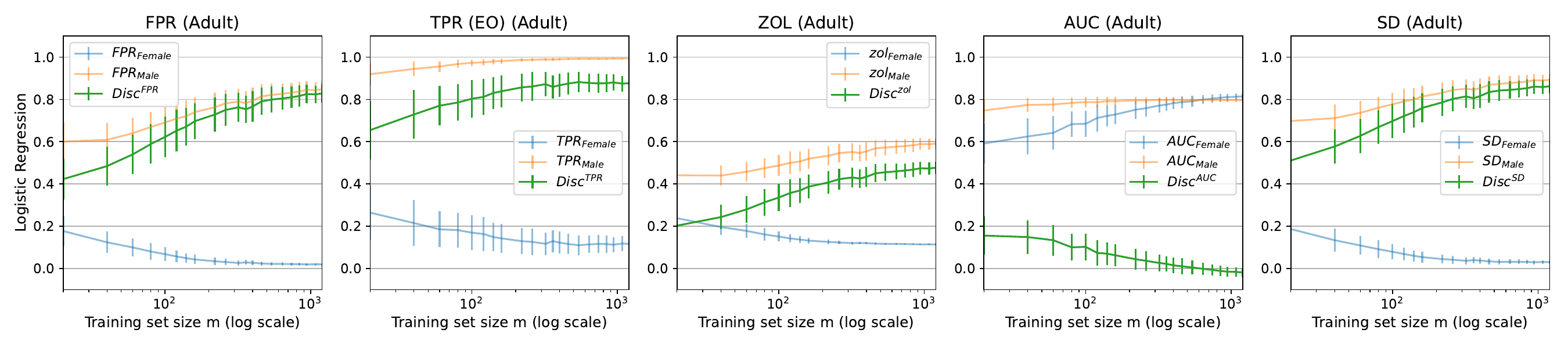}
    \caption{Behavior of discrimination metrics (along with protected group values) while increasing the training data size sampled from very discriminative population (postive outcome is 0.9 for privileged group and 0.1 for unprivileged group).}
    \label{fig:SSB_disc_LG_biased}
\end{figure}

\paragraph{\bf \textit{[Obs2]  \label{Obs2}{\em SSB} when training data is imbalanced}}Figure~\ref{fig:SSB_disc_LG_biased} shows the result of tracking the metrics values for each protected group and the corresponding discrimination behavior\footnote{Notice that discrimination value is the difference between the metric values of the privileged and the unprivileged groups.} for the Adult dataset (Results with the other datasets are provided in Appendix~\ref{sec:a_ssb}, Figure~\ref{fig:SSB_disc_LG_biased_all}). Surprisingly, training models using smaller data samples lead to less discrimination (the only exception is with {\em AUC} in the case of Adult dataset). This challenges the commonly accepted idea that training a model using limited size dataset amplifies discrimination.

\begin{figure}
    \centering
    \includegraphics[scale=0.32]{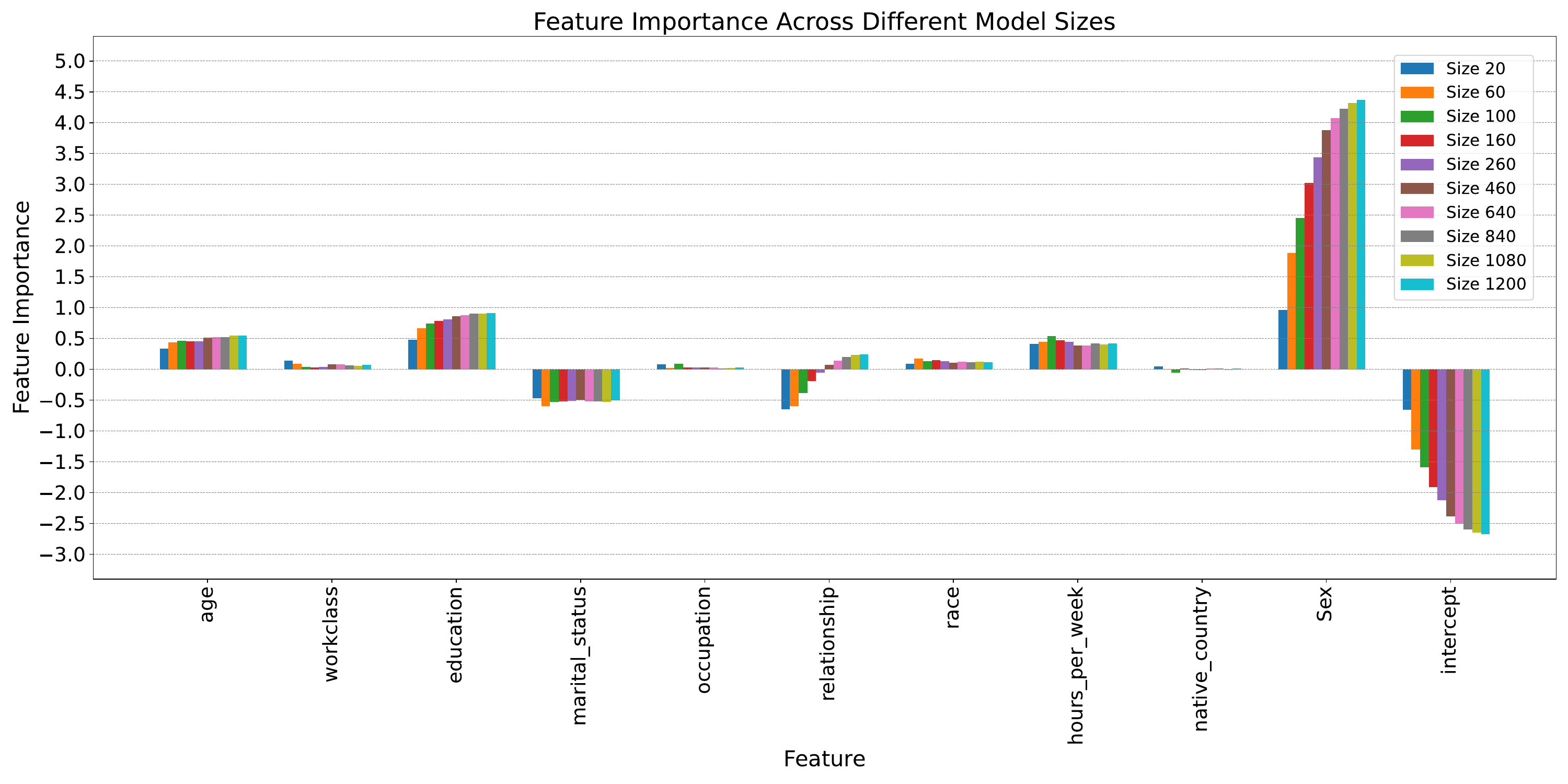}
    \caption{Importance of the sensitive attribute (Sex in Adult dataset) while increasing the data samples used for training.}
    \label{fig:sensAttrImportance1}
\end{figure}

\paragraph{\bf \textit{[Obs3] \bf \label{Obs3}{\em SSB} and feature importance}} The above counterintuitive behavior can be explained by observing how the relative importance of the sensitive attribute (Sex in the case of Adult dataset) changes while the model is trained with larger data samples. Figure~\ref{fig:sensAttrImportance1} shows that for small training data samples, the contribution of the sensitive attribute to the model is not significant, which translates into relatively low discrimination. As the training data sample is getting larger, the sensitive attribute contributes further to the model, and consequently, discrimination is higher. 

\begin{figure}
\small
    \centering
    \includegraphics[scale=0.4]{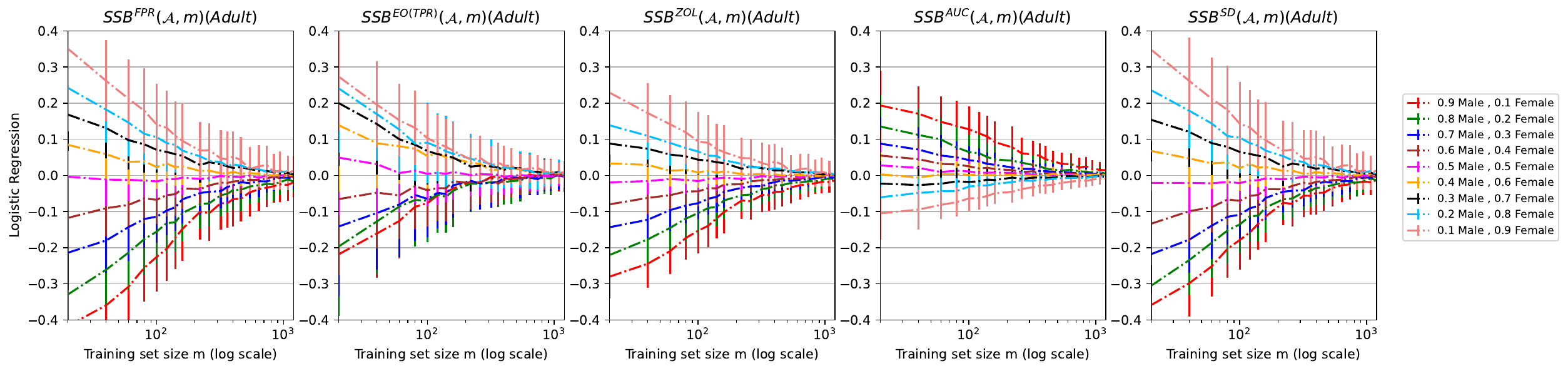}   
    \caption{Magnitude of {\em SSB} with varying level of imbalance and increasing size of the training data.}
    \label{fig:ImpactDataTypeOnSSB}
\end{figure}

\paragraph{\bf \textit{[Obs4] \bf \label{Obs4}Magnitude and effect of {\em SSB} depend on initial discrimination}} As mentioned above, Figures~\ref{fig:SSB_disc_LG_biased} and~\ref{fig:sensAttrImportance1} are obtained while training models using datasets featuring significant imbalance between protected groups. To understand better the effect of bias on discrimination, we consider further versions of datasets with varying level of imbalance ($0.8$ positive rate for privileged group while only $0.2$ for the unprivileged group, $0.7$ vs $0.3$, etc.) and with increasing size of the training dataset. The result of the experiment (Figure~\ref{fig:ImpactDataTypeOnSSB}) indicates that the more training data is imbalanced, the more significant the {\em SSB} will be while training model with smaller datasets. Hence the combined conclusion out of Figures~\ref{fig:SSB_disc_LG_biased} and ~\ref{fig:ImpactDataTypeOnSSB} is that: when the population features significant discrimination between protected groups, training a model using a small training data sample leads to significant bias ({\em SSB}) in estimating accurately the discrimination. However, the bias does not exacerbate the discrimination estimation, but on the contrary, it attenuates its value.

\begin{figure}
\small
    \centering
\includegraphics[scale=0.37]{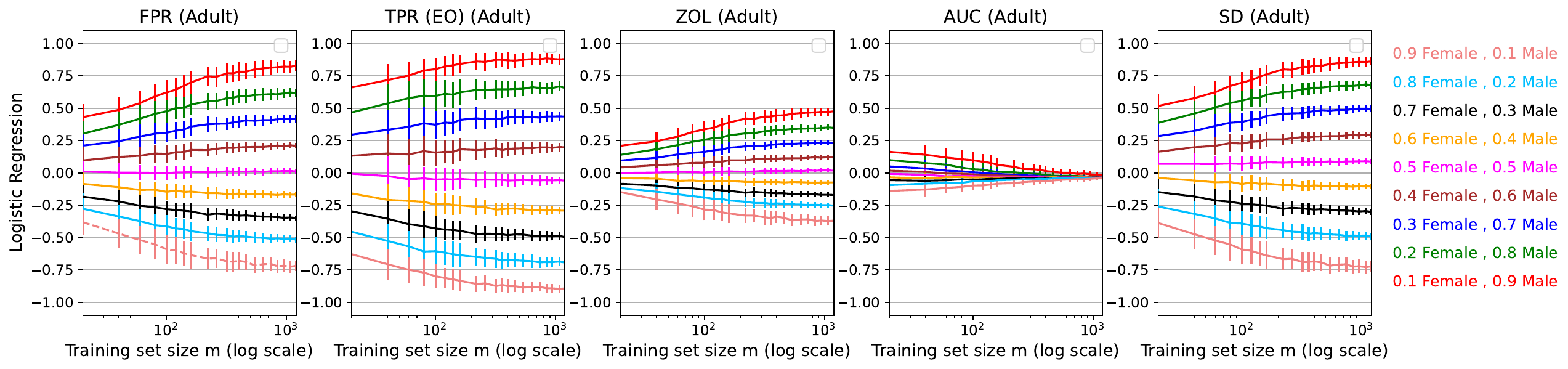}
    \caption{Magnitude of discrimination on increasingly imbalaned training data while increasing the training data size. 
    }
    \label{fig:Disc_increasing_imbalance}
\end{figure}

\paragraph{\bf \textit{[Obs5] \bf \label{Obs5}Magnitude of discrimination for increasingly imbalanced training data}} Figure~\ref{fig:Disc_increasing_imbalance} shows the behavior of discrimination values (not {\em SSB}) for different degrees of imbalance ($0.9$ vs $0.1$, $0.8$ vs $0.2$, $0.7$ vs $0.3$, etc.) while learning model based on increasingly large training datasets. One can notice that, the more discrimination (imbalance) there is in the population, the more significant the effect of {\em SSB} is on the estimated discrimination value. For instance, the bias due to {\em SSB} maybe as high as $0.38$ if the imbalance in the population is as high as $0.9$ vs $0.1$. Whereas, that same bias is almost absent ($\sim 0$) if no imbalance is observed in the population ($0.5$ vs $0.5$). Finally, notice how metrics that combine sensitivity and recall ({\em ZOL} and {\em AUC}) are less sensitive to the size of the training data than the remaining metrics.








\begin{figure}
    \hspace*{-2cm}
    \includegraphics[scale=0.55]{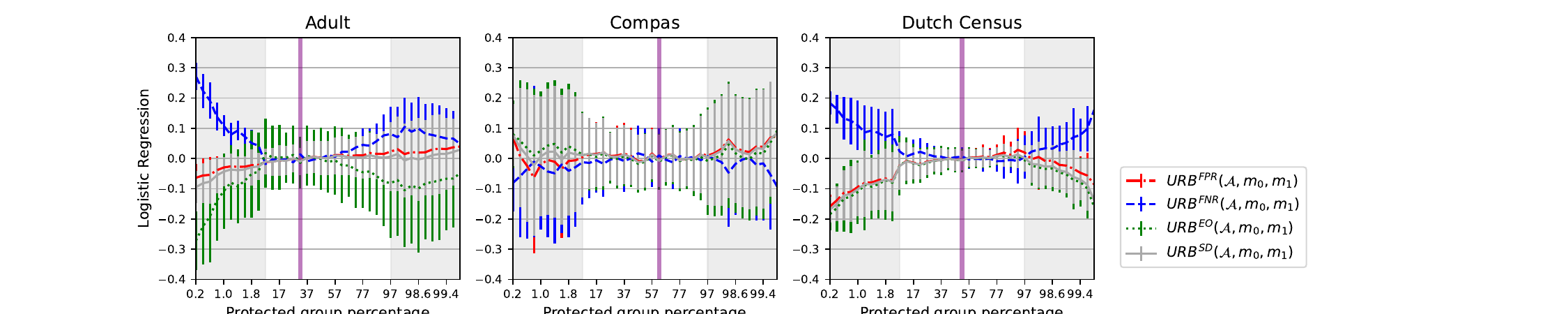}
    \hspace*{-2cm}
    \includegraphics[scale=0.55]{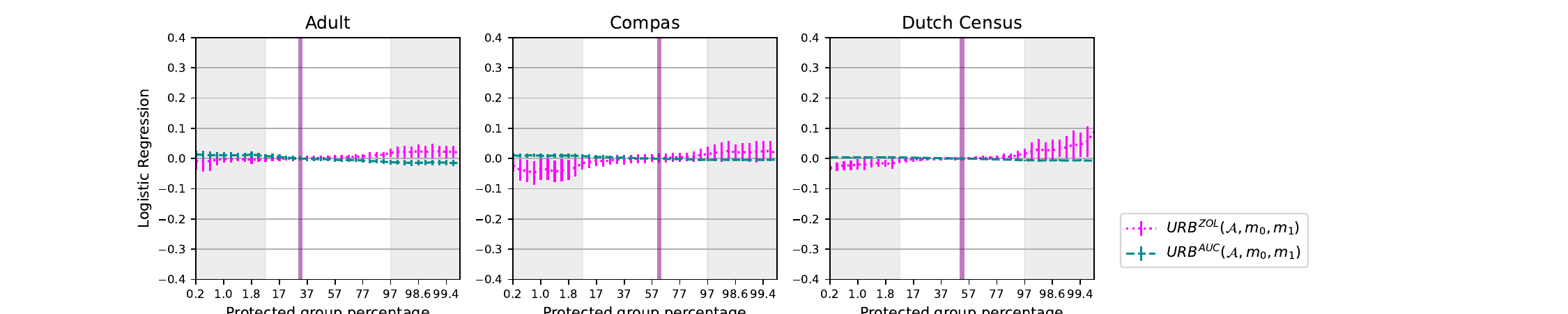}
    \caption{Underrepresentation Bias ({\em URB}) for different ratios of sensitive groups. The training set size is fixed ($1000$). The vertical bar represents the same ratio as the population. The shaded sections indicate a focus on the extreme proportions (less than $2\%$ and more than $98\%$).}
    \label{fig:urb_lg_orig}
\end{figure}


\subsection{Underrepresentation Bias ({\em URB})}
\label{sec:exp_urb}

 \paragraph{\bf \textit{[Obs6] \bf \label{Obs6}Magnitude of underrepresentation bias ({\em URB})}} Figure~\ref{fig:urb_lg_orig} shows the behavior of {\em URB} as the proportion of the sensitive group increases for the same three datasets and for using logistic regression as learning algorithm. The purple vertical bar indicates the percentage of the sensitive group in the entire dataset (population). For instance, for the adult dataset, the percentage of females is $31\%$. The shaded parts in the background of Figure~\ref{fig:urb_lg_orig}'s plots indicate that we are ``zooming'' on the extreme values (the plots are using different steps for the shaded and unshaded parts\footnote{The step is very small below $2\%$ and above $98\%$.}). Almost all plots exhibit the same pattern for {\em URB}, that is, the further the proportions of sensitive groups are from the population proportions reference (vertical bar), the higher is the bias. The same expected behavior for {\em URB} is obtained when using the other classifiers  (Figures~\ref{fig:all_URB1} and~\ref{fig:all_URB2} in Appendix~\ref{sec:a_urb}). The resilience of {\em AUC} and {\em ZOL} metrics to extreme training set sizes holds also for imbalanced training sets. Notice that $URB^{AUC}$ and $URB^{ZOL}$ remain stable even for extremely imbalanced training sets.



\begin{figure}
    \centering
    \hspace{-2cm}
    \includegraphics[scale=0.4]{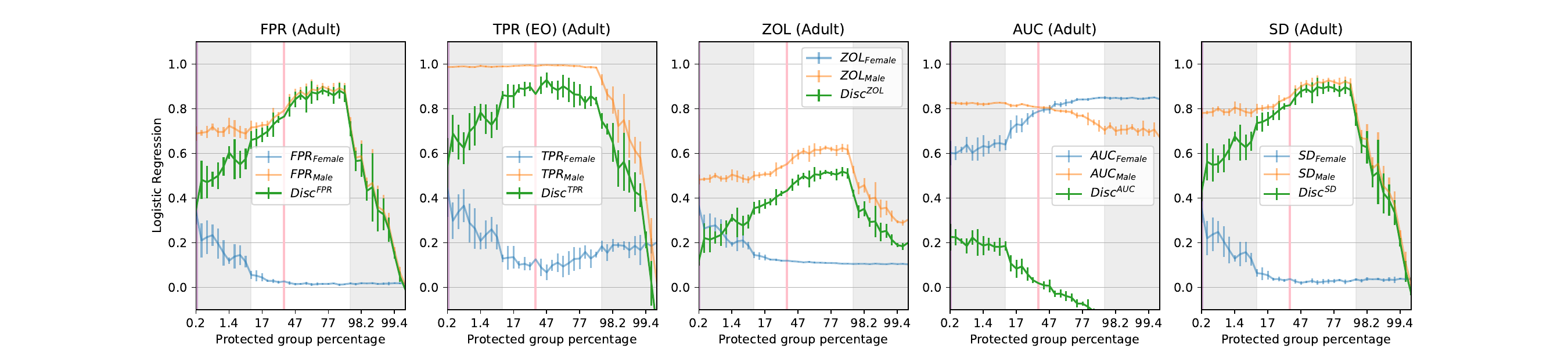}
    \caption{{\em URB} Experiment using Logistic Regression as the Training Algorithm with extreme bias (0.9 vs 0.1).}
    \label{fig:URB_veryBiased}
\end{figure}

\paragraph{ \bf \textit{[Obs7] \bf \label{Obs7}Magnitude of discrimination in {\em URB} experiment}}Figure~\ref{fig:URB_veryBiased} shows the discrimination value (green solid line) obtained as the difference between the two other values (orange and blue lines). For most of the metrics, discrimination is higher for more balanced splitting values ($0.5$, etc.) and hence lower for extremely imbalanced splitting values ($0.01$, etc.). Since we know from the previous experiment (Figure~\ref{fig:urb_lg_orig}) that {\em URB} is higher for extreme splitting values, we can deduce that {\em URB} has an attenuating effect on the initially high discrimination. 

\begin{figure}
\small
	{\includegraphics[scale=0.38]{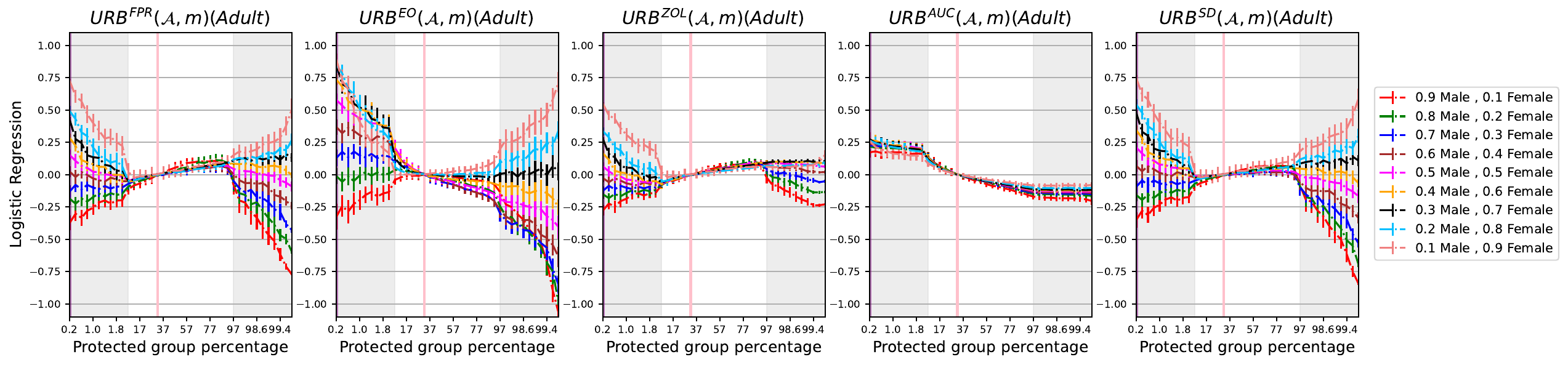}}
	 \caption{Magnitude of {\em URB} when training data is increasingly discriminative.}
	\label{fig:URB_initialBias}
\end{figure}

\paragraph{\bf \textit{[Obs8] \bf \label{Obs8}Magnitude of {\em URB} is proportional to the initial discrimination}} Figure~\ref{fig:urb_lg_orig} shows that the magnitude of {\em URB} is proportional to the extremeness of the splitting values. The following experiment shows that the magnitude of {\em URB} depends also on the initial discrimination in the data. Figure~\ref{fig:URB_initialBias} depicts the magnitude of {\em URB} when considering versions of the data with an increasing level of initial discrimination. One can notice that for the same splitting value, {\em URB} is higher when the data is intially more discriminating. The plots confirm also that the metrics that combine sensitivity and recall are, again, less sensitive to {\em URB} than the remaining metrics.

\begin{figure}
\small
 \includegraphics[scale=0.38]{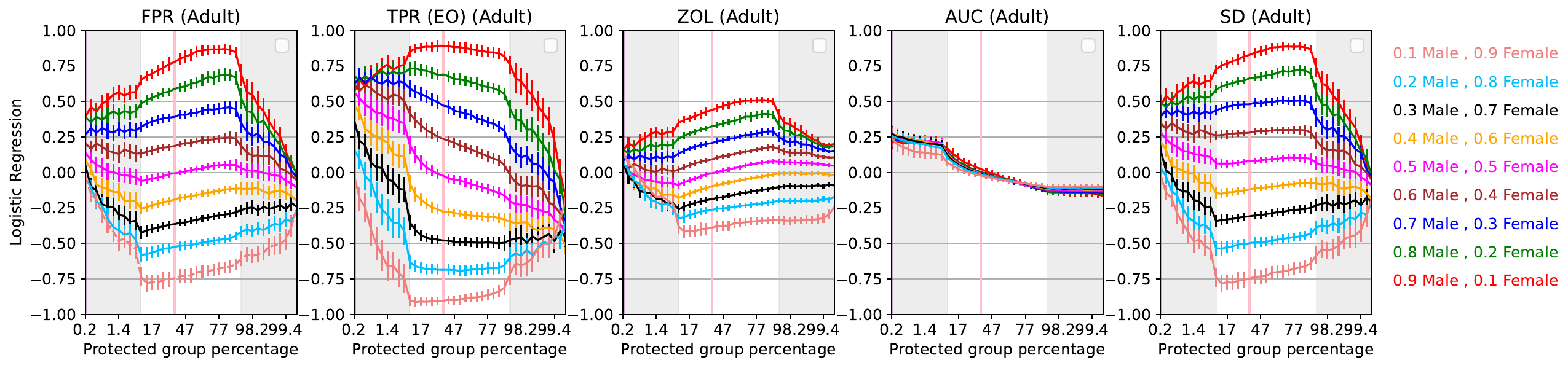}	
	 \caption{Magnitude of discrimination on increasingly imbalanced training data while changing the proportions of protected groups.}
	\label{fig:Disc_urb_initial}
\end{figure}

\paragraph{\bf \textit{[Obs9] \bf \label{Obs9}Magnitude of discrimination for increasingly imbalanced training data}} Building on the above observations about the behavior of {\em URB}, Figure~\ref{fig:Disc_urb_initial} shows when {\em URB} exacerbates discrimination, when it attenuates it, and the magnitude of the effect. As a summary, when data features initial significant discrimination between the protected groups (e.g. red color lines), discrimination is much lower for extreme splitting values (both left and right sides of the plots) compared to balanced splitting values (middle parts of the plots). This disparity can be explained by the attenuating effect of {\em URB}. On the contrary when data is relatively fair (no disparity between the positive outcome of protected groups, e.g. pink color lines), discrimination is higher for extreme splitting values (grayed parts of the plots) compared to more balanced splitting values (middle parts of the plots). This disparity can be explained by the exacerbating effect of {\em URB}. Finally, in terms of the importance of the effect of {\em URB}, one can notice that it is proportional to the initial discrimination in the data (e.g. the effect is more significant for the red color lines than for the green color lines).

\subsection{Bias Mitigation}
\label{sec:mitig}

\begin{figure}
\centering
    \hspace{-2cm}
    \includegraphics[scale=0.4]{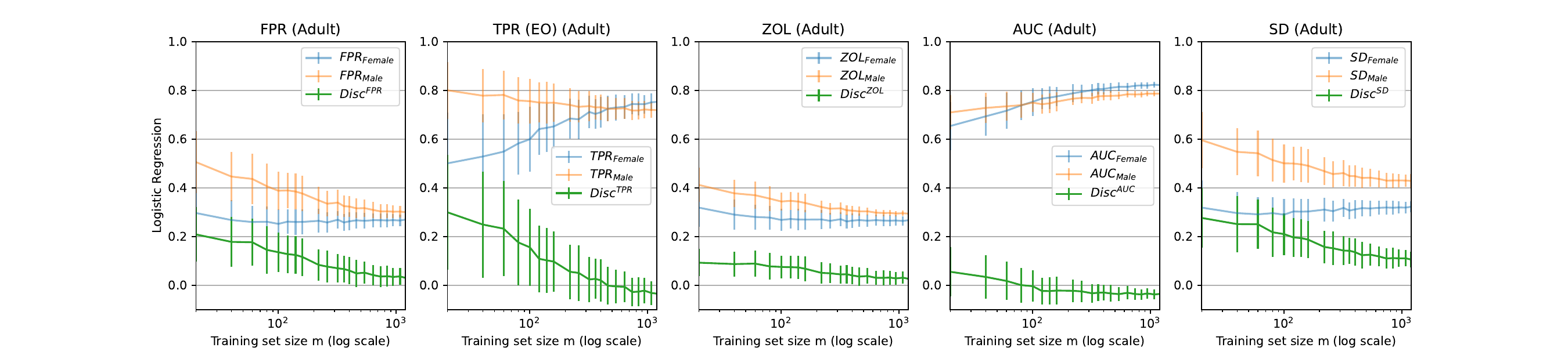}
    \caption{Effect of Bias Mitigation using Reweighing Preprocessing on Discrimination in Logistic Regression for {\em SSB} Experiment (increasing the training data size)}
    \label{fig:ReweighingDiscSSB}
\end{figure}

  \paragraph{\bf \textit{[Obs10] Effect of Reweighing pre-processing on {\em SSB}}} Figure~\ref{fig:ReweighingDiscSSB} shows the discrimination (green color lines) while training the model using an increasingly large training dataset. It is easy to see that according to all metrics, discrimination is decreasing and converging to $0$ as models are trained using larger datasets. This is strikingly different than Figure~\ref{fig:SSB_disc_LG_biased} where discrimination is gradually revealed as training data size increases. Hence, the important observation out of Figure~\ref{fig:ReweighingDiscSSB} is that to see a significant impact of Reweighing in mitigating discrimination, a relatively large training dataset should be used. The effect of Reweighing is diluted when the training dataset is small. 

\begin{figure}
\centering
    \hspace{-2cm}
    \includegraphics[scale=0.4]{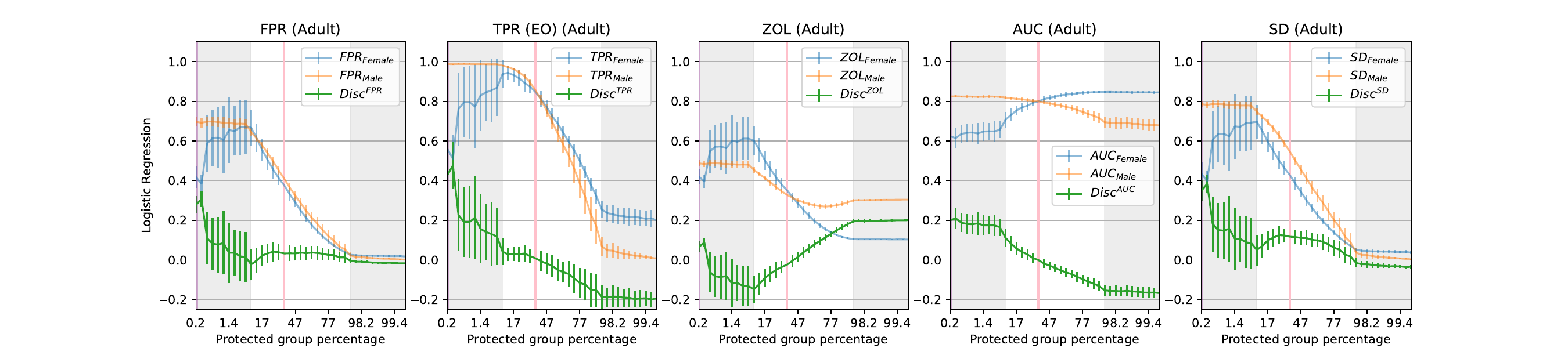}
    \caption{Effect of Bias Mitigation using Reweighing Preprocessing on Discrimination in Logistic Regression for {\em URB} Experiment}
    \label{fig:ReweighingDiscURB}
\end{figure}

\paragraph{\bf \textit{[Obs11] \label{Obs11} Effect of Reweighing pre-processing on {\em URB}}} Using the same significantly unfair version of the benchmark datasets, the next experiment aims at observing the effect of Reweighing when data features a disparity in representation between the protected groups ({\em URB}). Figure~\ref{fig:ReweighingDiscURB} shows that Reweighing is efficient at mitigating discrimination in presence of {\em URB} as discrimination (green color line) is close to $0$ and does not increase significantly with extreme over/under-representation (grayed areas).  

\begin{figure}
    \centering
    \hspace{-2cm}
    \includegraphics[scale=0.4]{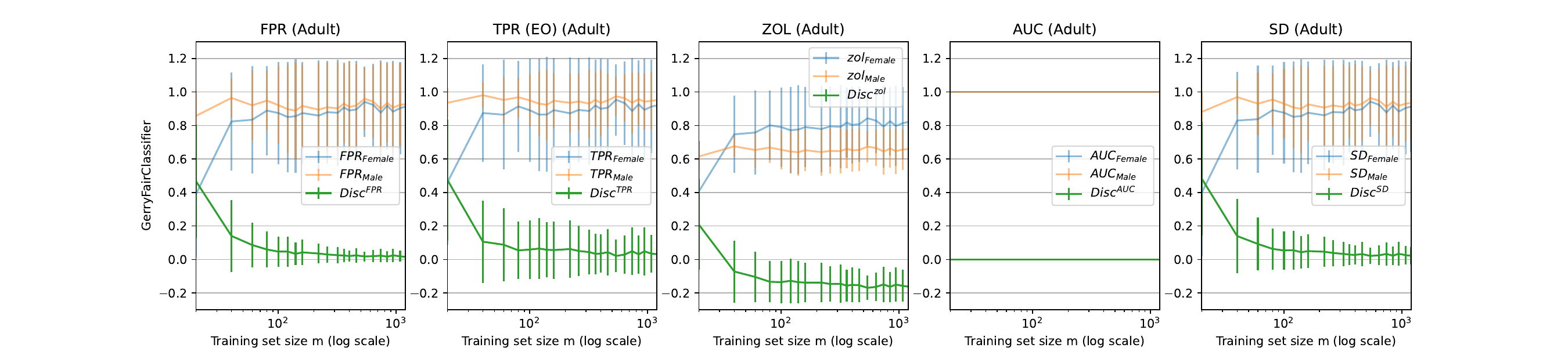}
    \caption{Effect of bias mitigation using GerryFairClassifier in-processing on discrimination for {\em SSB} experiment}
    \label{fig:GerryDiscSSB}
\end{figure}

\begin{figure}[h]
\centering
    \hspace{-2cm}
    \includegraphics[scale=0.4]{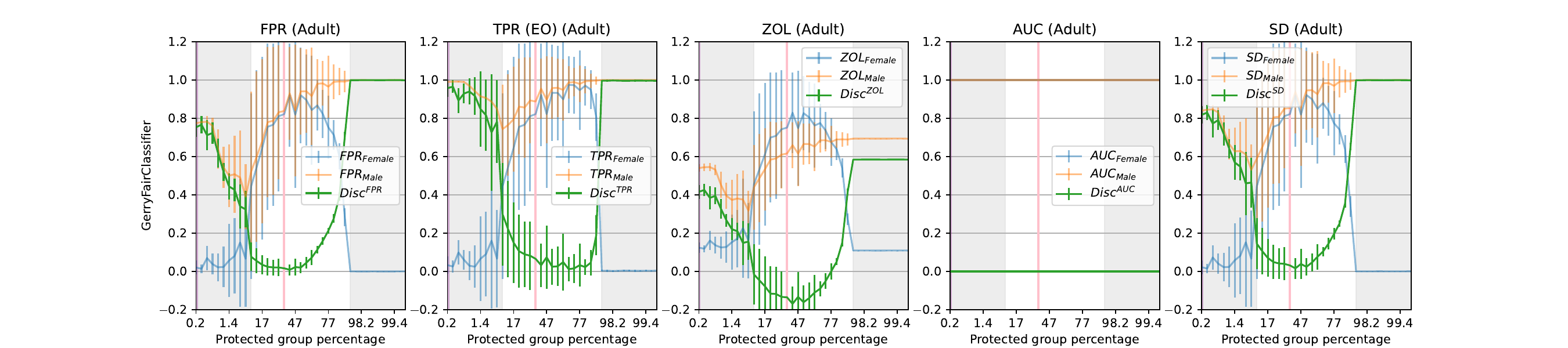}
    \caption{Effect of bias mitigation using GerryFairClassifier in-processing on discrimination for {\em URB} experiment.}
    \label{fig:GerryDiscURB}
\end{figure}

\paragraph{\bf \textit{[Obs12] \label{Obs12}Effect of GerryFairClassifier in-processing on {\em SSB} and {\em URB}}} Figures~\ref{fig:GerryDiscSSB} and~\ref{fig:GerryDiscURB} show the results of the same experiment but using the GerryFairClassifier in-processing mitigation approach. While increasing the training dataset produces the same discrimination behavior as Reweighing (Figure~\ref{fig:ReweighingDiscSSB}), under/over representation of protected groups revealed different behavior than Reweighing (Figure~\ref{fig:ReweighingDiscURB}). GerryFairClassifier approach mitigates bias efficiently only in the absence of under or over representation of protected groups (Figure~\ref{fig:GerryDiscURB}). Notice that with extreme underrepresentation (grayed areas), GerryFairClassifier fails to mitigate discrimination. On the opposite, it exacerbates it. 


\subsection{Effect of collecting more samples on discrimination}
\label{sec:exp_threshold}

\begin{figure}[h]
\small
    \centering
    \hspace*{-2cm}
     \includegraphics[height=1.4in, width=7in]{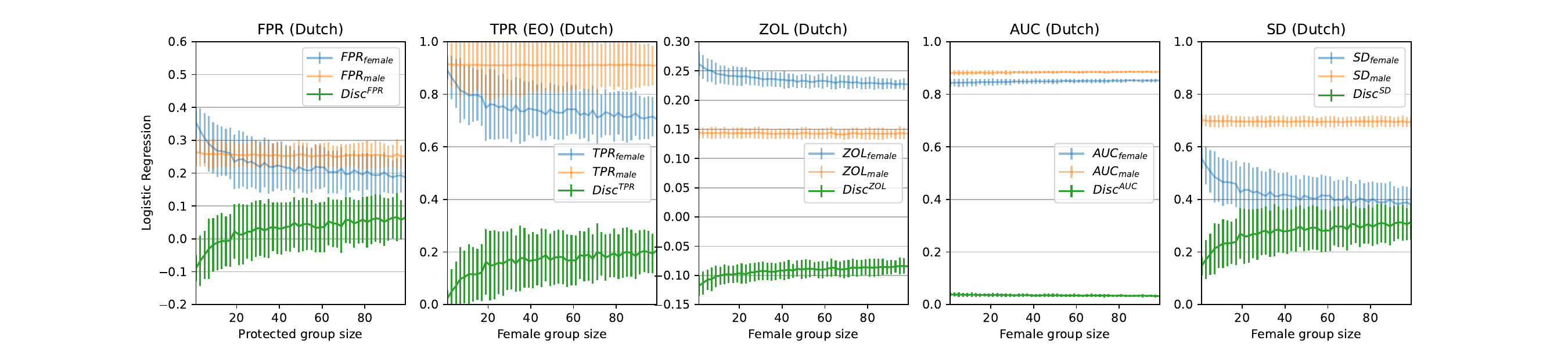}
    \caption{Discrimination while augmenting the training set with female group samples randomly. The male group size is fixed at $100$. Dataset is Dutch Census and training algorithm is logistic regression.}
    \label{fig:DutchThreshold}
\end{figure}


\begin{figure}[h]
\small
    \centering
    \hspace*{-2cm}
    \includegraphics[height=1.4in, width=7in]{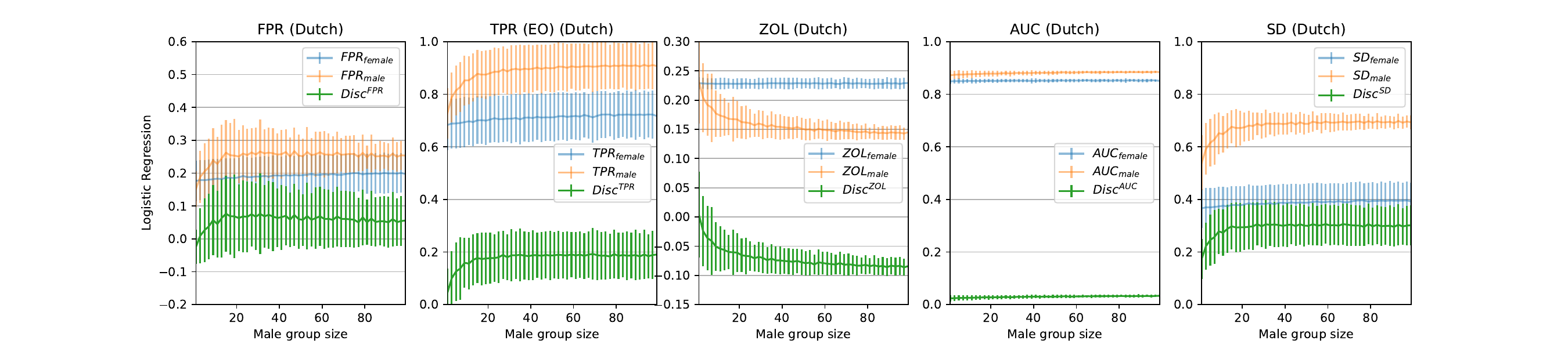}
    \caption{Discrimination while augmenting the training set with male group samples randomly. The female group size is fixed at $100$. Dataset is Dutch Census and training algorithm is logistic regression.}
    \label{fig:DutchThreshold_inv}
\end{figure}

\paragraph{\bf \textit{[Obs13] \label{Obs13}Random data augmentation}}
Figure~\ref{fig:DutchThreshold} shows how the metrics values for each group, as well as the corresponding difference (discrimination) are changing as more protected group samples are considered for model training. As expected, the metric values for the male group maintain the same mean while for the female group, it is changing. Interestingly, according to all metrics (except AUC), discrimination is increasing as data is more balanced. Figure~\ref{fig:DutchThreshold} shows the results with logistic regression, but the pattern is similar for other classification algorithms (Figure~\ref{fig:DutchThreshold_all} in Appendix~\ref{sec:a_threshold}) and for other benchmark datasets (Figure~\ref{fig:adultThreshold_all} in Appendix~\ref{sec:a_threshold}). This counterintuitive behavior is also observed for the reverse experiment where the protected group (female) sample size is fixed ($100$ samples) while the privileged group (male) is under-represented and more samples are collected and considered in the training (Figure~\ref{fig:DutchThreshold_inv}). It is important to mention that in all previous experiments, selecting samples to balance the training set is performed randomly to simulate, as accurately as possible, data collection in real scenarios. 
The fairness enhancing potential of adding more samples for the sensitive group depends on the initial fairness characteristics of the data and the goal of the classifier. Wang et al.~\cite{Wang2018BalancedDA} point out that adding more samples of the minority group to the data increases predictive accuracy and fairness specifically in the classification tasks, where sensitive attribute is part of the output of classification, for example face recognition~\cite{buolamwini2018gender}. 

\begin{figure}[h]
\small
    \centering
    \hspace*{-2cm}
    \includegraphics[height=1.4in, width=7.2in]{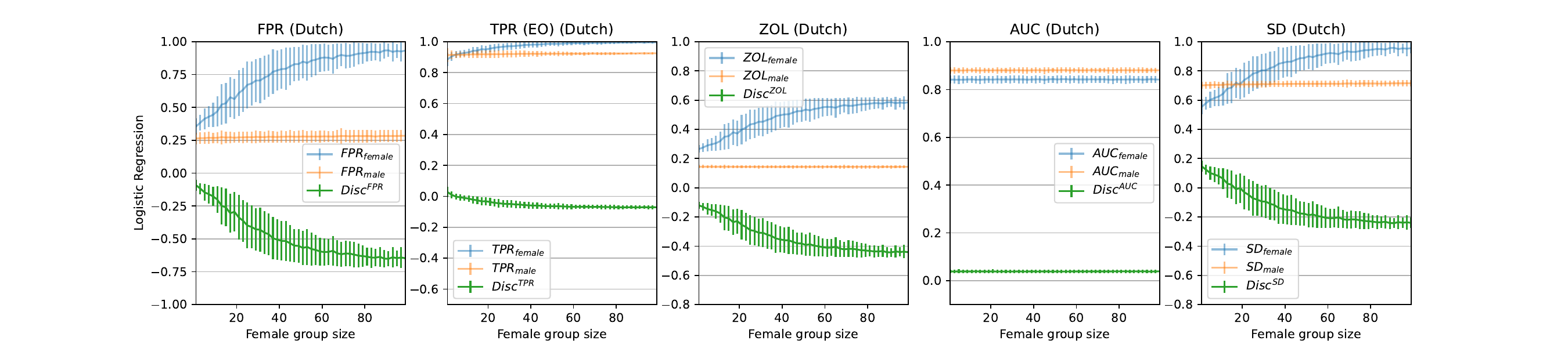}
    \caption{Discrimination while augmenting the training set with only positive outcome female group samples. The male group size is fixed at $100$. Dataset is Dutch Census and training algorithm is logistic regression.}
    \label{fig:DutchThreshold_sel}
\end{figure}

\paragraph{\bf \textit{[Obs14] \label{Obs14}Selective data augmentation}}
If, however, training set is balanced by selecting a specific type of samples, in particular, protected group samples with positive outcome, discrimination will be decreasing as data gets balanced (Figure~\ref{fig:DutchThreshold_sel}). In all three experiments (collecting more protected group samples randomly, collecting more unprotected group samples randomly, and collecting only positive outcome protected group samples), the importance of the sensitive feature (Sex) in the prediction (\textit{shap} explanation~\cite{shap}) behaves the same way (Figure~\ref{fig:sensImportance} in Appendix~\ref{sec:a_threshold}), that is, it contributes more to the learned model as the data is more balanced.

\section{Recommendations for Practitioners}
Based on the observations obtained from Section~\ref{sec:exp}, one can propose
the following recommendations for a practitioner considering fair model learning but cannot guarantee that the training data is free from any type of imbalance. 



\paragraph{\em \bf A. When training data is small, metrics that combine sensitivity and specificity are more reliable.} When training data is of limited size, there will be a significant bias due to sample size limitation ({\em SSB}). The recommendation is to use fairness metrics that combine sensitivity and specificity (e.g. AUC, ZOL, F1, etc.) as they are less sensitive to that type of bias (\textbf{\em Obs1,} \autoref{Obs1}).

\paragraph{\em \bf B. With biased data, a small size training data attenuates discrimination rather than exacerbates it.} When training data is suspected to be significantly biased, a model trained using a limited-size dataset may underestimate the level of discrimination. This may look contradicting the previous recommendation but this can be explained using the magnitude of {\em SSB} (\textbf{\em Obs2}, \autoref{Obs2} and \textbf{\em Obs3}, \autoref{Obs3}).

\paragraph{\em \bf C. More initial discrimination leads to more biased estimation of discrimination.} The effect of the size of training data on the accuracy of the estimation is proportional to the degree of  imbalance in the training data (\textbf{\em Obs4}, \autoref{Obs4} and \textbf{\em Obs5}, \autoref{Obs5}).

\paragraph{\em \bf D. Under representation of a group leads to inaccurate estimation of discrimination} When the training data features under-representation of one or several sensitive groups, common fairness metrics lead to a significant bias in measuring discrimination. The recommendation is again to use metrics that combine sensitivity and specificity, such as AUC and ZOL, which are less sensitive to under-representation (\textbf{\em Obs6}, \autoref{Obs6}).

\paragraph{\em \bf E. The effect of underrepresentation of a group may under or over-estimate discrimination depending on the initial discrimination in the training data.} The magnitude of underrepresentation bias ({\em URB}) depends on the extremeness of the disparity between sensitive group sizes. However, the type of the effect of bias on discrimination (attenuation or exacerbation) depends on the initial discrimination in the training data (\textbf{\em Obs7}, \autoref{Obs7},  \textbf{\em Obs8}, \autoref{Obs8} and \textbf{\em Obs9}, \autoref{Obs9}).

\paragraph{\em \bf F. The efficiency of bias mitigation is diminished by small training data and disparity in group representation.} Whether for pre-processing or in-processing, bias mitigation is only efficient when training data is large enough and sensitive groups are well represented. For small training data or data featuring underrepresented sensitive groups, common mitigation approaches fail to address discrimination appropriately (\textbf{\em Obs10}, 
\textbf{\em Obs11}, \autoref{Obs11} and \textbf{\em Obs12}, \autoref{Obs12}).

\paragraph{\em \bf E. Random data augmentation can amplify discrimination} Random data augmentation can unexpectedly amplify discrimination. In our experiments, adding random samples from the underrepresented group often overestimated discrimination  (\textbf{\em Obs13}, \autoref{Obs13} and \textbf{\em Obs14}, \autoref{Obs14}). To explain this phenomenon we distinguish two types of prediction tasks, that lead to different fairness mitigation strategies. When the sensitive attribute is relevant to the prediction (e.g., face recognition or rare disease detection), collecting more minority group data helps mitigate bias. However, for tasks where the sensitive attribute should not influence predictions (e.g., college admissions or loan decisions, as in our experiments), adding data from the same biased distribution reinforces the unfair link between the sensitive attribute and the outcome. Balancing data with respect to the outcome, rather than oversampling the minority class, is a more effective solution.

\section{Conclusion}

A common source of bias in machine learning arises from limited or imbalanced training data. This paper introduces $SSB$ and $URB$ to systematically capture these biases. Through empirical analysis on benchmark datasets using standard classification algorithms, we expose relevant observations on sampling bias.
First, discrimination metrics, such as equalized odds, which account for the trade-off between precision and recall (e.g. based on $AUC$, $ZOL$, etc.) are more robust to sampling biases compared to other more common metrics (e.g. based on $FPR$, equal opportunity, etc.). Therefore, when working with limited or imbalanced training data, fairness metrics such as equalized odds, which incorporate precision-recall trade-offs, are recommended for reliable discrimination measurement.
Second, in presence of significant discrimination in the population, training a model using a limited size or imbalanced dataset underestimates discrimination rather than overestimating it.
Third, the effect of typical bias mitigation approaches (pre-processing and in-processing) is diluted by the limited size and/or the imbalance of the training dataset.
Fourth, collecting more samples from an underrepresented group can amplify discrimination when drawn from the same biased distribution. This highlights the need for balanced data augmentation strategies that account for both the imbalance in the data and the type of classification task, rather than merely oversampling the minority class. By addressing these biases systematically, this paper underscores the importance and provides recommendations for tailored mitigation strategies and robust fairness metrics to ensure reliable and equitable machine learning outcomes.


\bibliographystyle{abbrv}
\bibliography{texz-bibFile}

\newpage
\appendix

\section{Appendix}

\subsection{Additional plots for the magnitude of SSB (Section~\ref{sec:ssbexp})}
\label{sec:a_ssb}
\begin{figure}[H]
    \centering    \includegraphics[height=8in, width=7.6in]{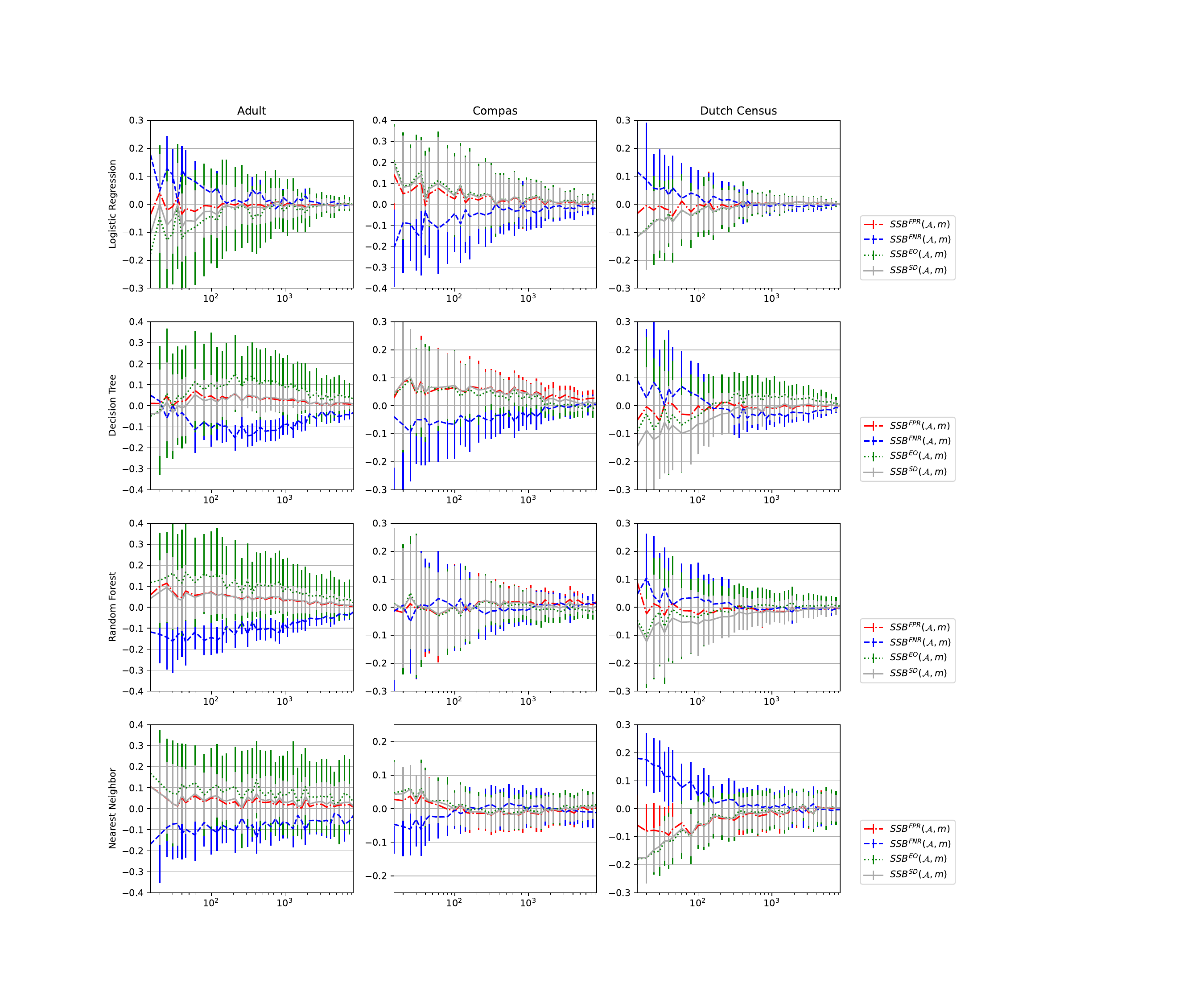}
    \caption{Magnitude of sample size bias (SSB) for increasing size of the training data.}
    \label{fig:all_ssb1}
\end{figure}

\begin{figure}[H]
    \centering
    \includegraphics[height=8in, width=7.6in]{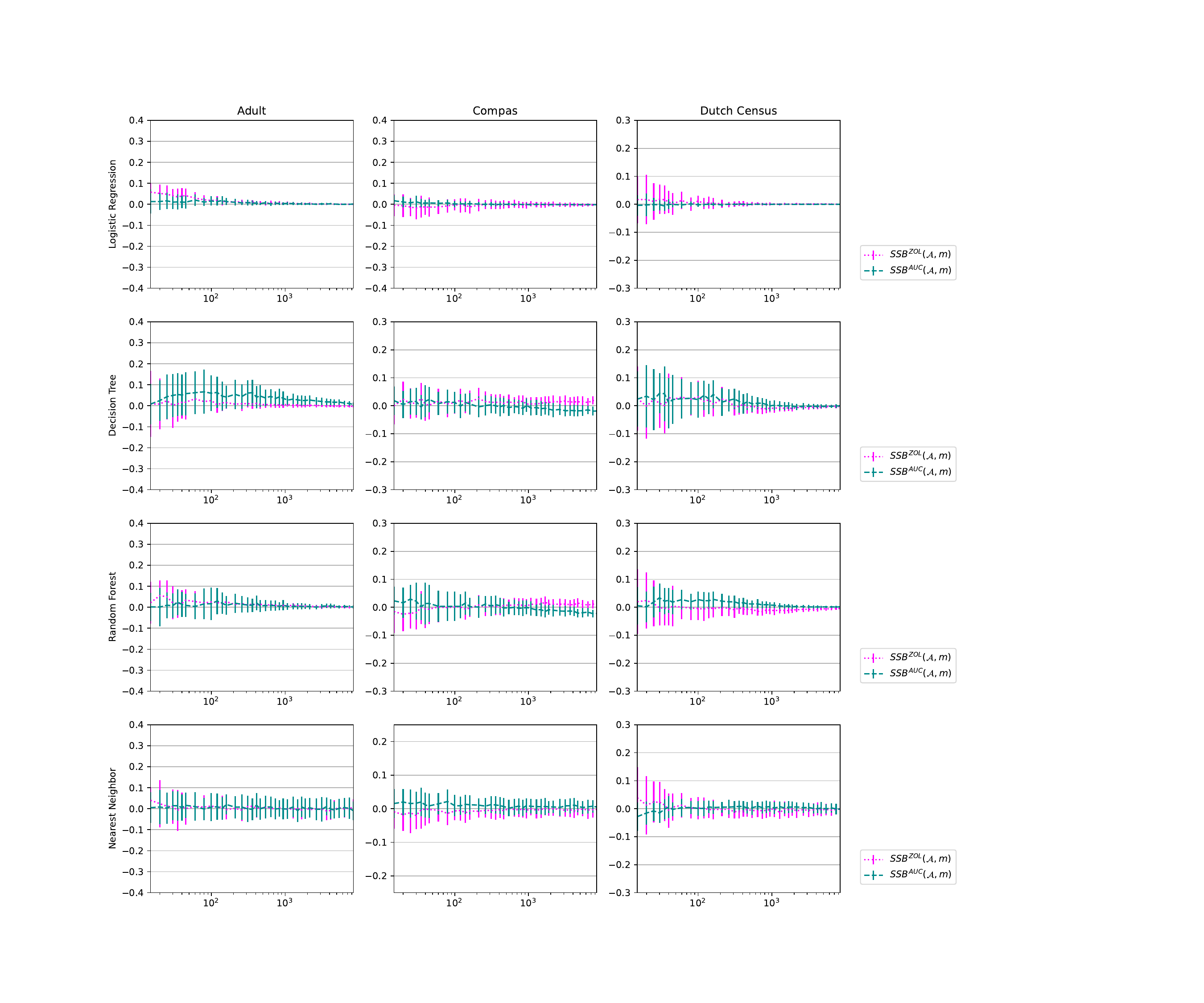}
    \caption{Magnitude of sample size bias (SSB) for increasing size of the training data.}
    \label{fig:all_ssb2}
\end{figure}

\begin{figure}[H]
    \centering
\includegraphics[scale=0.37]{Plots/SSB/THE_Last_Disc_New_Disparity_Experience_0.9_Positive_Male_0.1_Positive_Female_Adult_Logistic_Regression.pdf}
\includegraphics[scale=0.37]{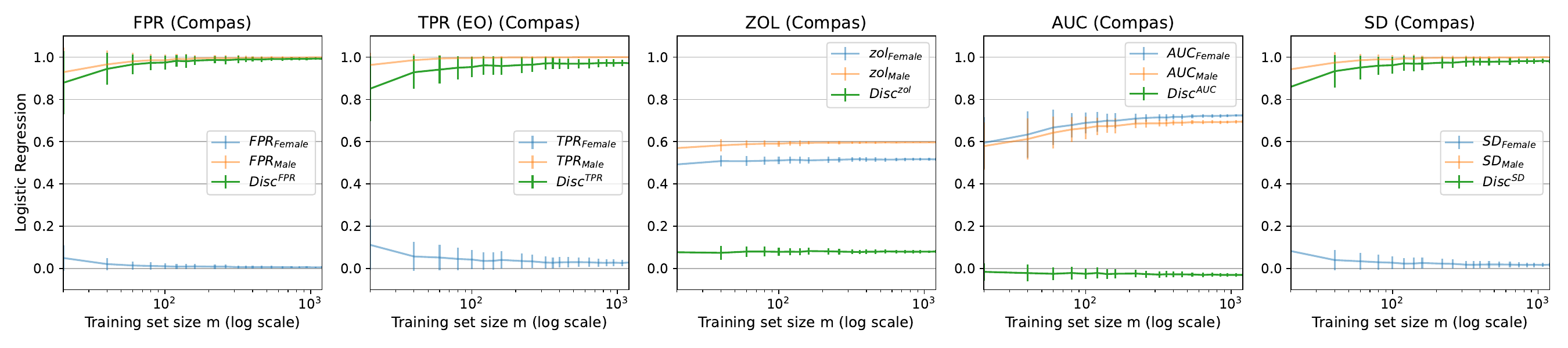}
\includegraphics[scale=0.37]{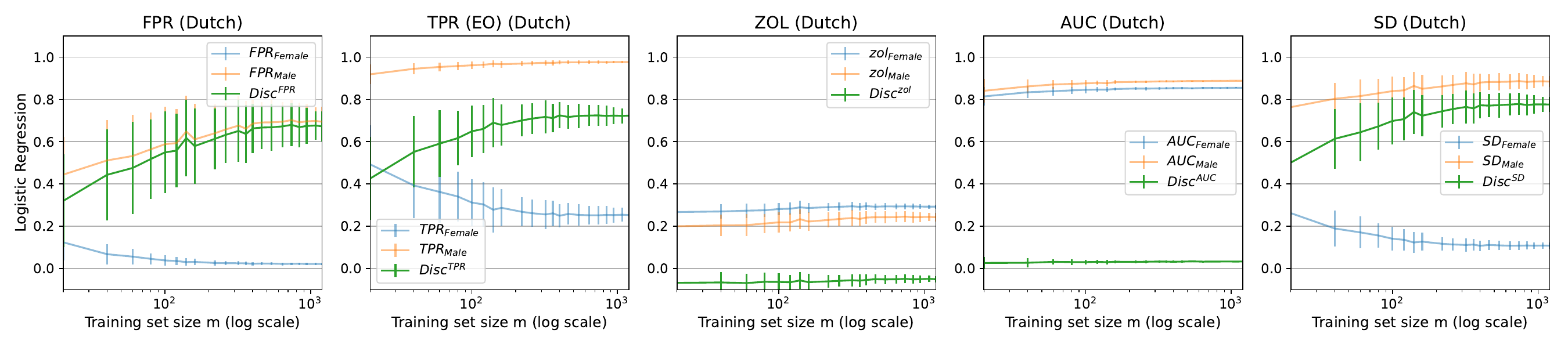}
    \caption{Behavior of discrimination metrics (along with protected group values) while increasing the training data size sampled from very discriminative population (postive outcome is 0.9 for privileged group and 0.1 for unprivileged group).}
    \label{fig:SSB_disc_LG_biased_all}
\end{figure}

\begin{figure}[H]
    \centering
    \includegraphics[scale=0.37]{Plots/SSB/SSB_Disparity_of_Label_for_sensitive_attribute_Logistic_Regression_Adult.pdf}
    
    \includegraphics[scale=0.37]{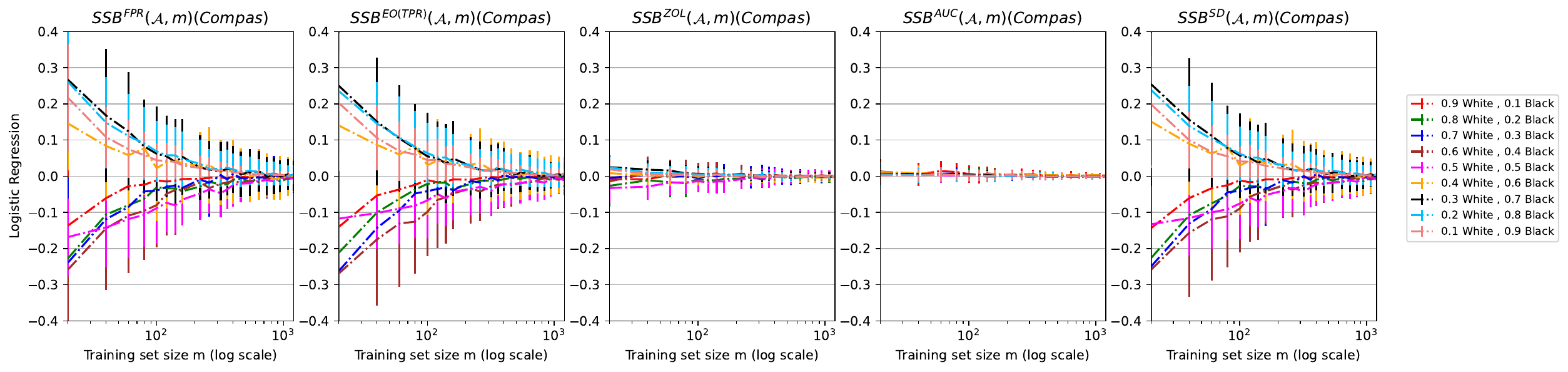}
    \includegraphics[scale=0.37]{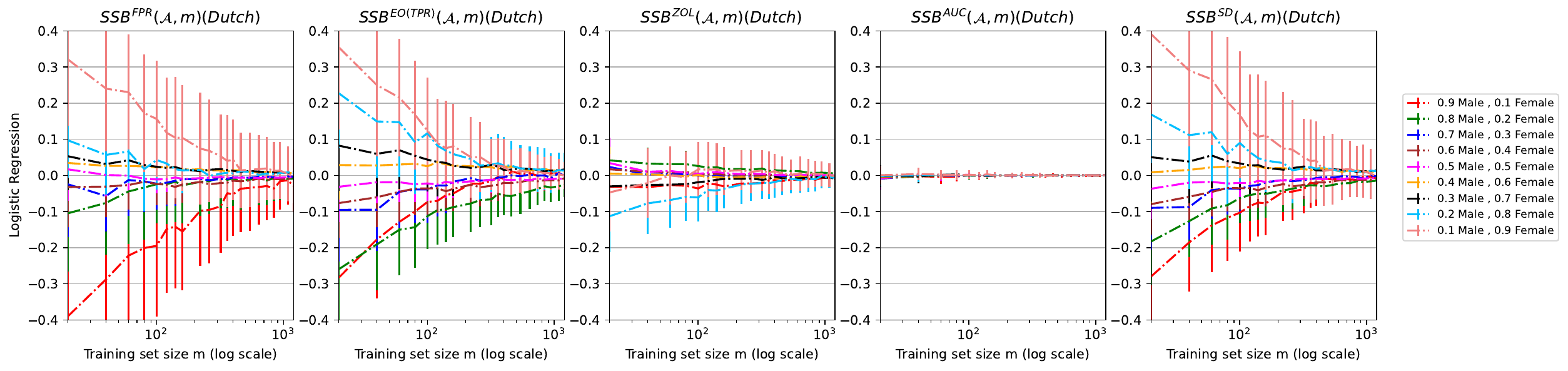}
    \caption{Magnitude of {\em SSB} with varying level of imbalance and increasing size of the training data.}

\end{figure}


\begin{figure}[H]
    \centering
    \includegraphics[scale=0.37]{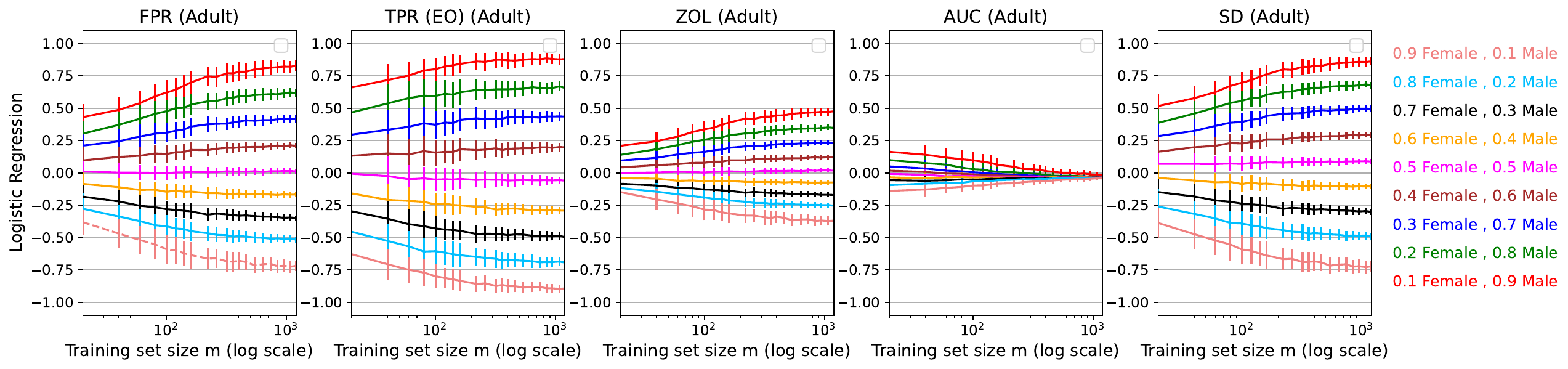}
    \includegraphics[scale=0.37]{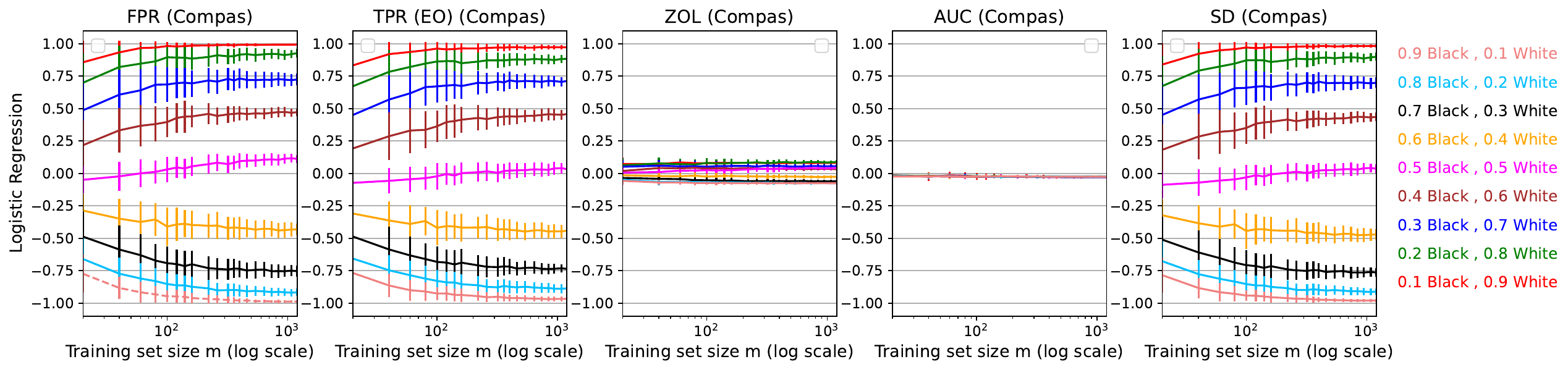}
    \includegraphics[scale=0.37]{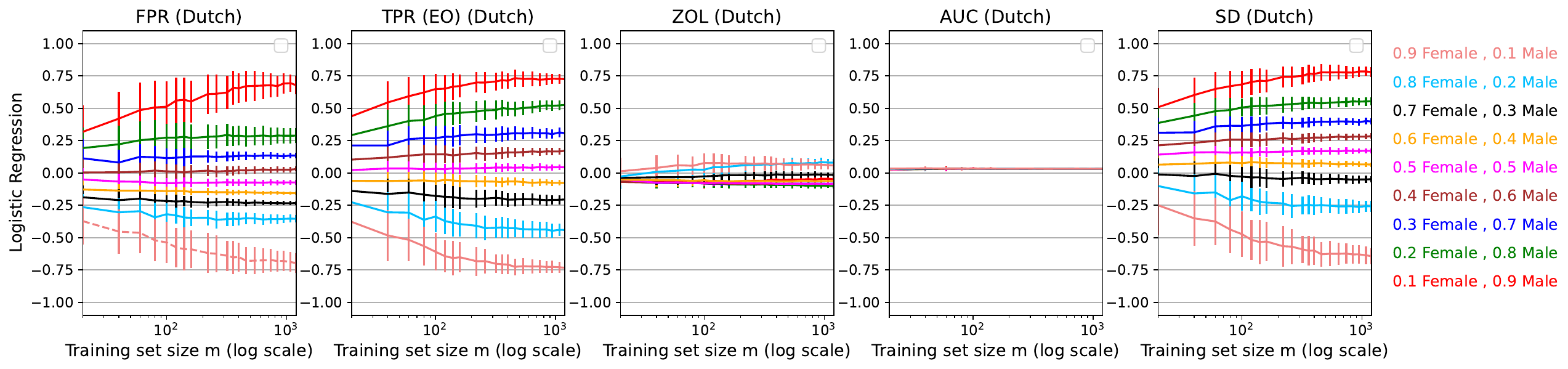}
    
    \caption{Magnitude of discrimination on increasingly imbalaned training data while increasing the training data size. 
    }

\end{figure}

\subsection{Additional plots for Magnitude of discrimination in {\em URB} experiment (Section~\ref{sec:exp_urb})}

 \begin{figure}[H]
    \centering
    \hspace*{-2cm}
    \includegraphics[height=8in, width=7.6in]{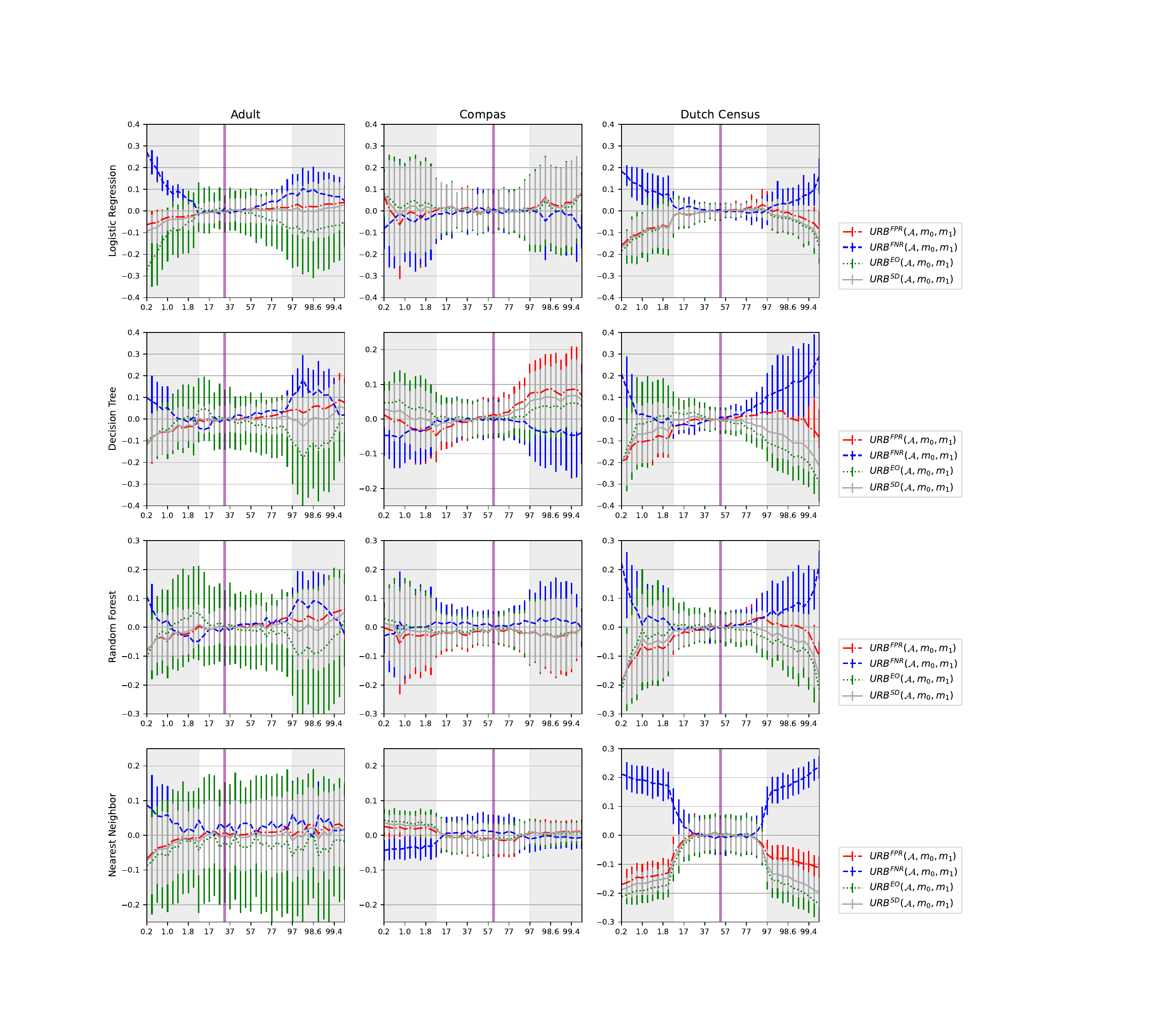}
    \caption{Underrepresentation Bias (URB) for different ratios of sensitive groups. The training set size is fixed ($1000$). The horizontal bar represents the same ratio as the population. The shaded sections indicate a focus on the extreme proportions (less than $2\%$ and more than $98\%$).}
    \label{fig:all_URB1}
\end{figure}

\begin{figure}[H]
    \centering
    \hspace*{-2cm}
    \includegraphics[height=8.2in, width=7.7in]{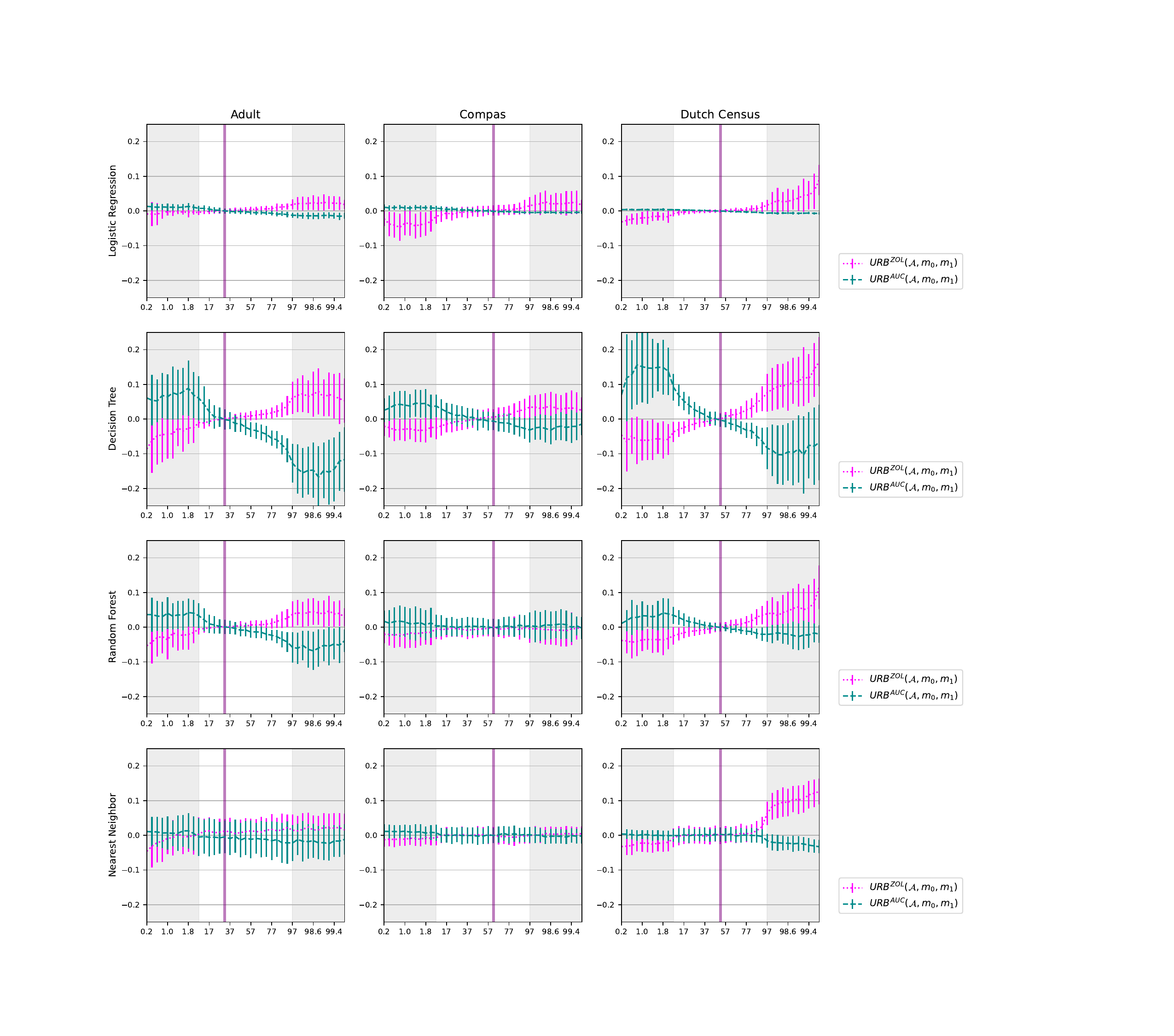}
    \caption{Underrepresentation Bias (URB) for different ratios of sensitive groups. The training set size is fixed ($1000$). The horizontal bar represents the same ratio as the population. The shaded sections indicate a focus on the extreme proportions (less than $2\%$ and more than $98\%$).}
    \label{fig:all_URB2}
\end{figure}

\begin{figure}[H]
    \centering
    \includegraphics[scale=0.35]{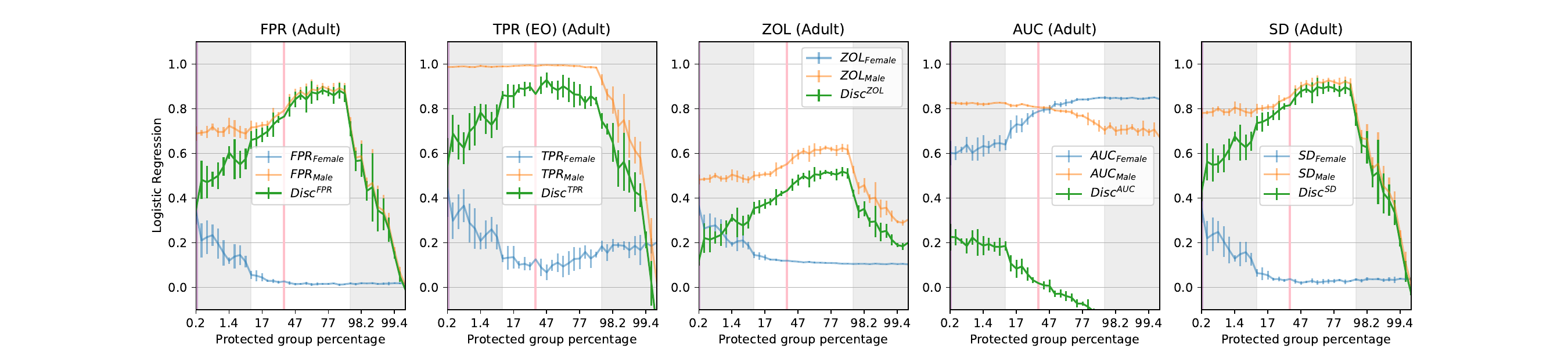}
    \includegraphics[scale=0.35]{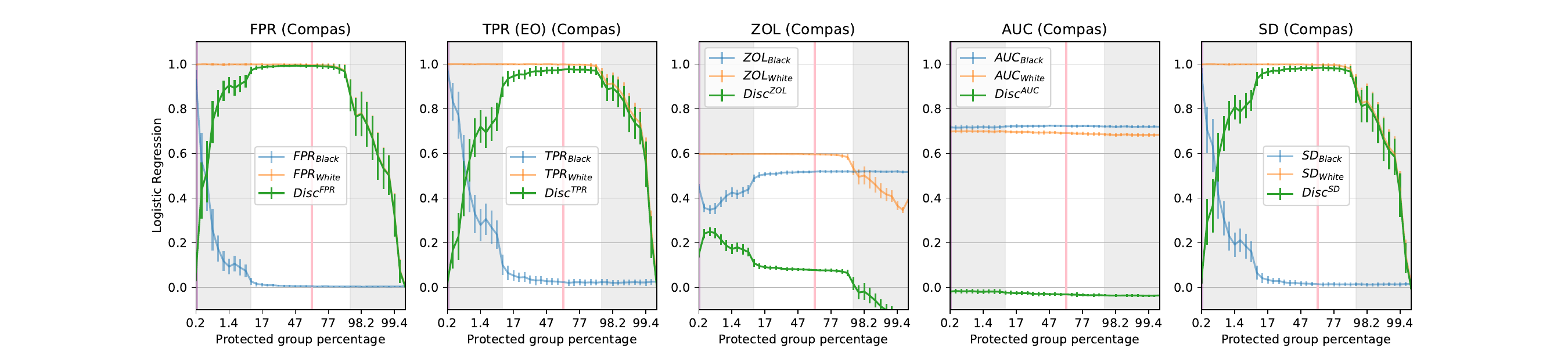}
    
    \includegraphics[scale=0.35]{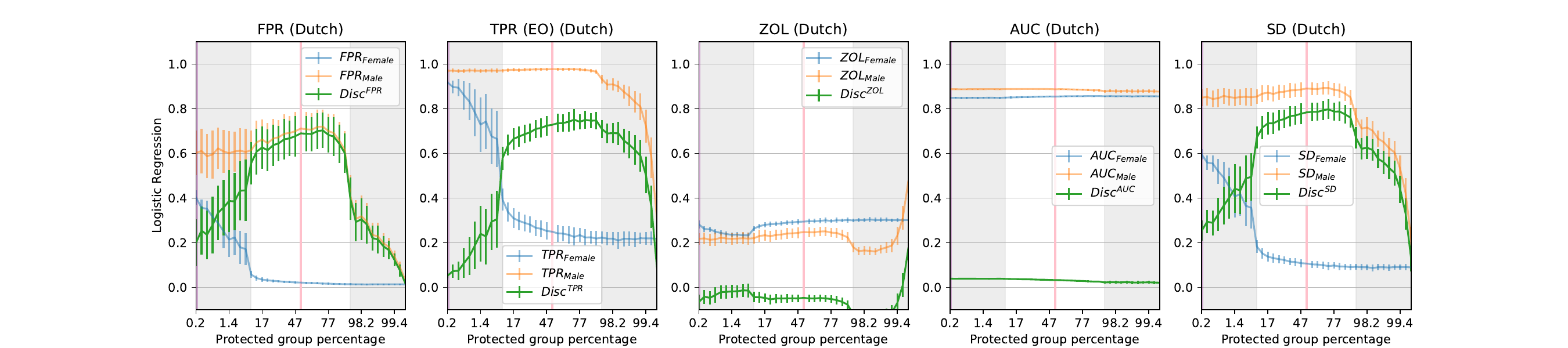}
    
        \caption{{\em URB} Experiment using Logistic Regression as the Training Algorithm with extreme bias (0.9 vs 0.1).}

\end{figure}


\begin{figure}[H]
    \centering
    \includegraphics[scale=0.35]{Plots/URB/URB_Disparity_of_Sensitive_LG_Adult.pdf}
    \includegraphics[scale=0.35]{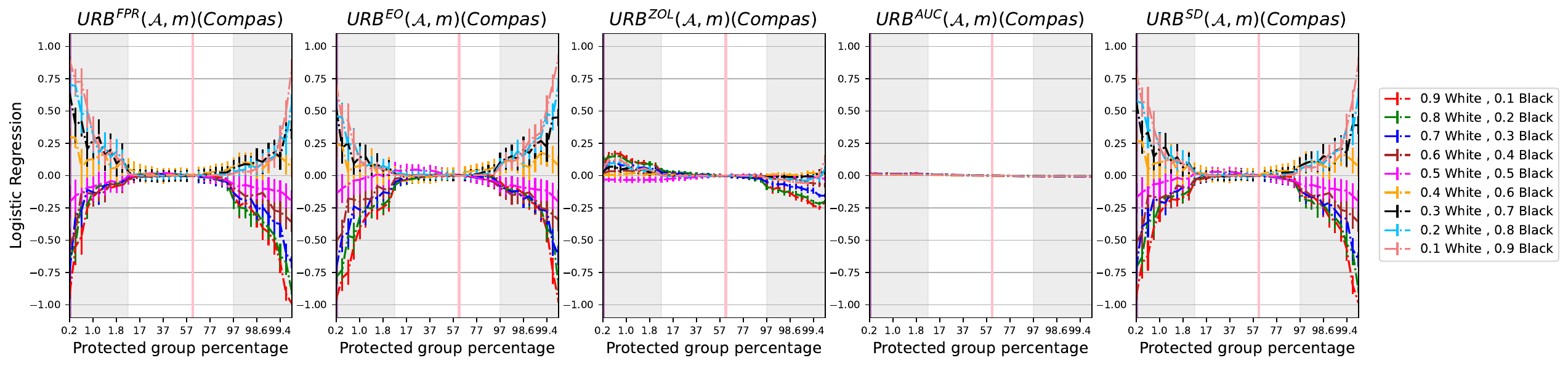}
    \includegraphics[scale=0.35]{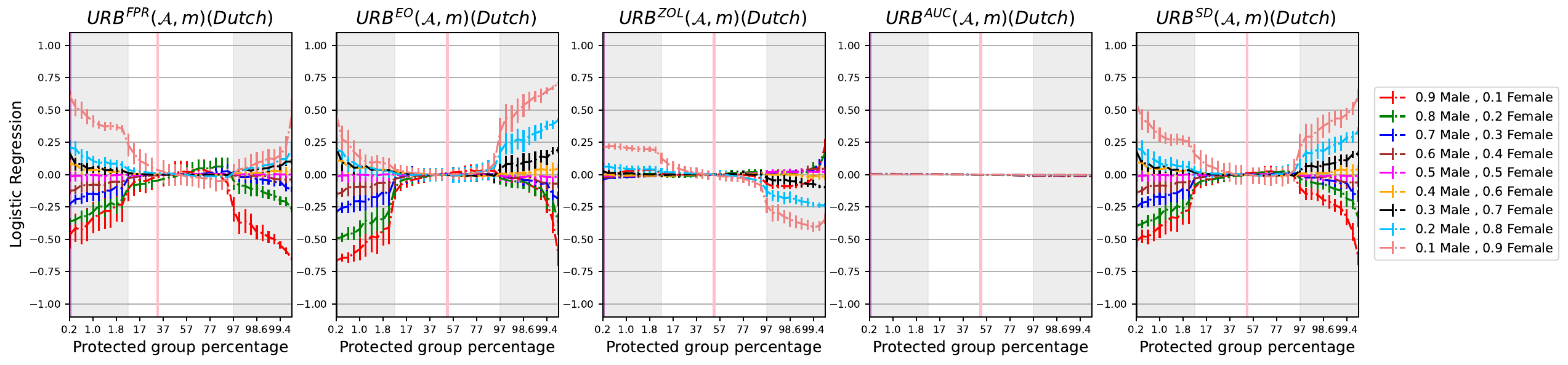}

	 \caption{Magnitude of {\em URB} when training data is increasingly discriminative.}

\end{figure}


\begin{figure}[H]
    \centering
    \includegraphics[scale=0.35]{Plots/URB/Disc_Disparity_of_Sensitive_URB_LG_Adult.pdf}
    \includegraphics[scale=0.35]{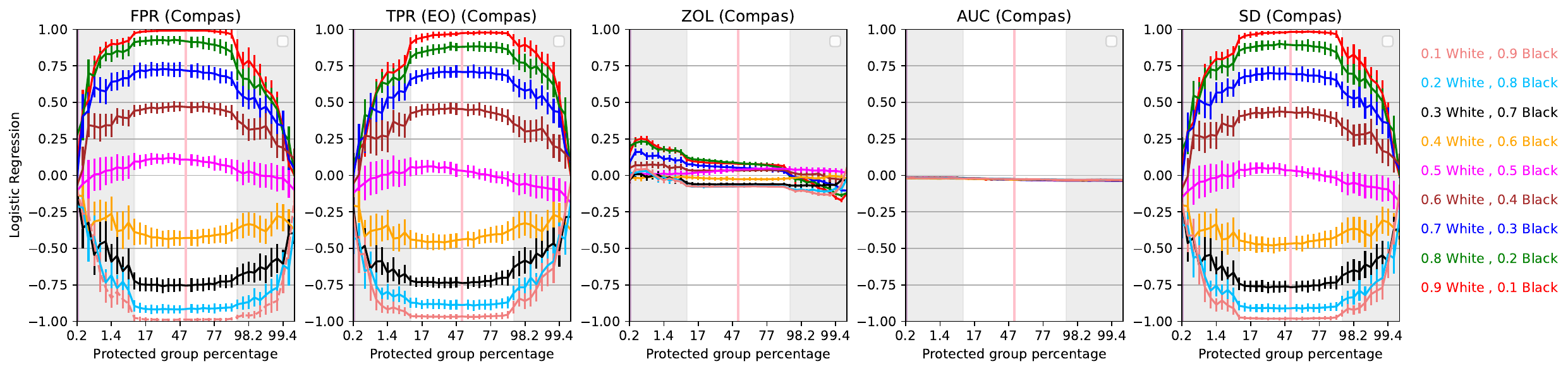}
    \includegraphics[scale=0.35]{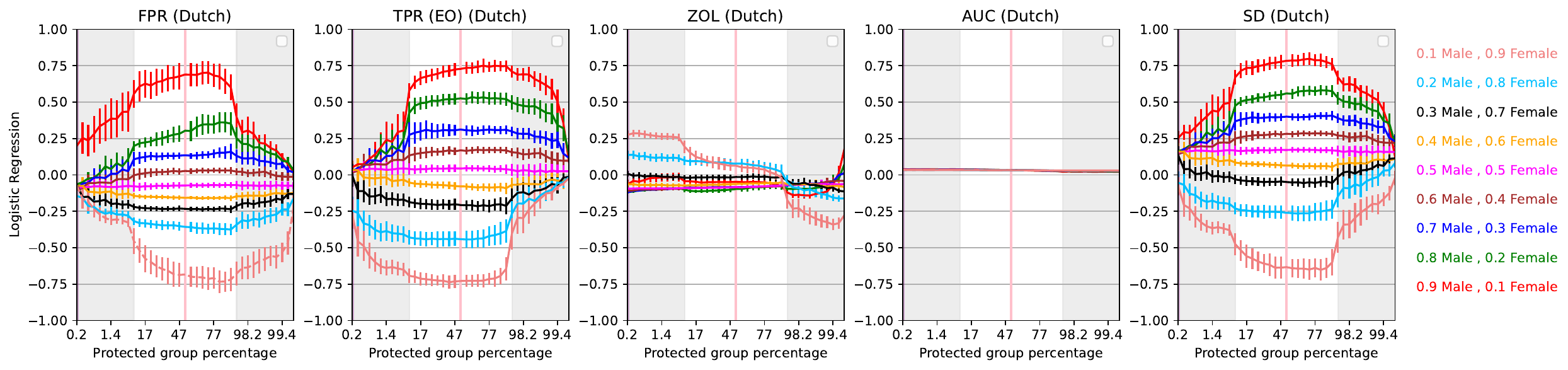}

        	 \caption{Magnitude of discrimination on increasingly imbalanced training data while changing the proportions of protected groups.}

\end{figure}

\subsection{Additional plots for the effect of bias mitigation (Section~\ref{sec:mitig})}

\begin{figure}[H]
    \centering
    \includegraphics[scale=0.38]{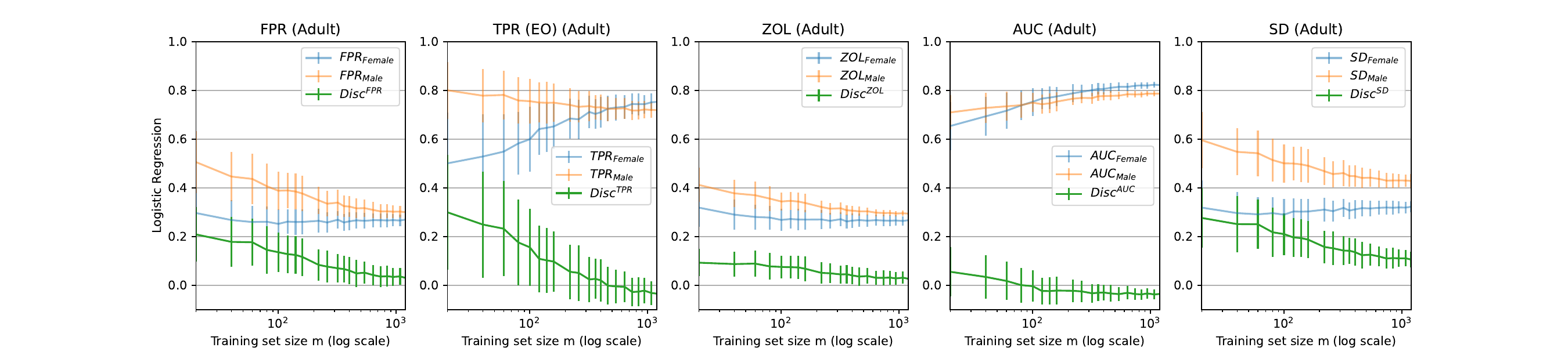}
    \includegraphics[scale=0.38]{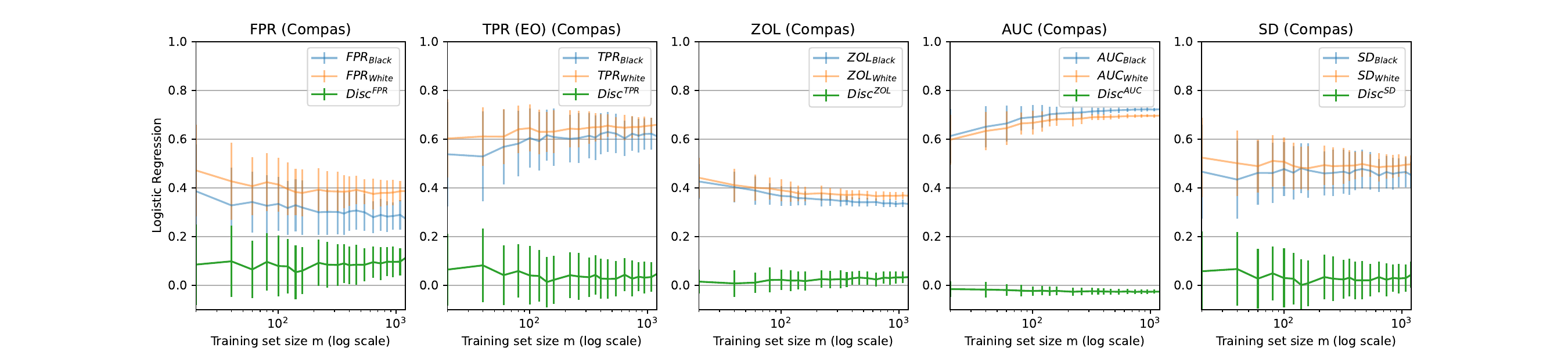}
    \includegraphics[scale=0.38]{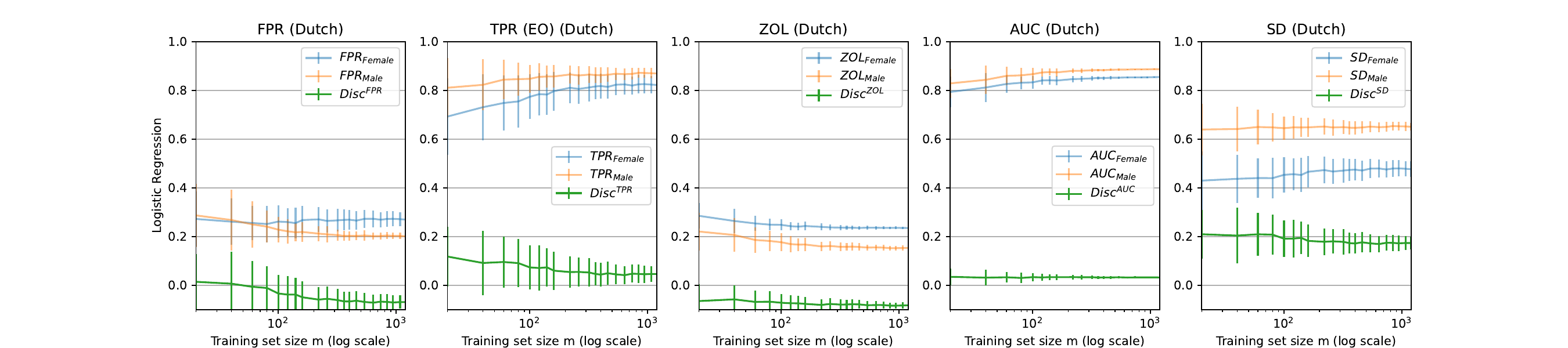}
    \caption{Effect of Bias Mitigation using Reweighing Preprocessing on Discrimination in Logistic Regression for {\em SSB} Experiment (increasing the training data size)}

\end{figure}


\begin{figure}[H]
    \centering
    \includegraphics[scale=0.38]{Plots/reweighing/disc_reweighing_urb_0.1_F_0.9_M_Adult.pdf}
    \includegraphics[scale=0.38]{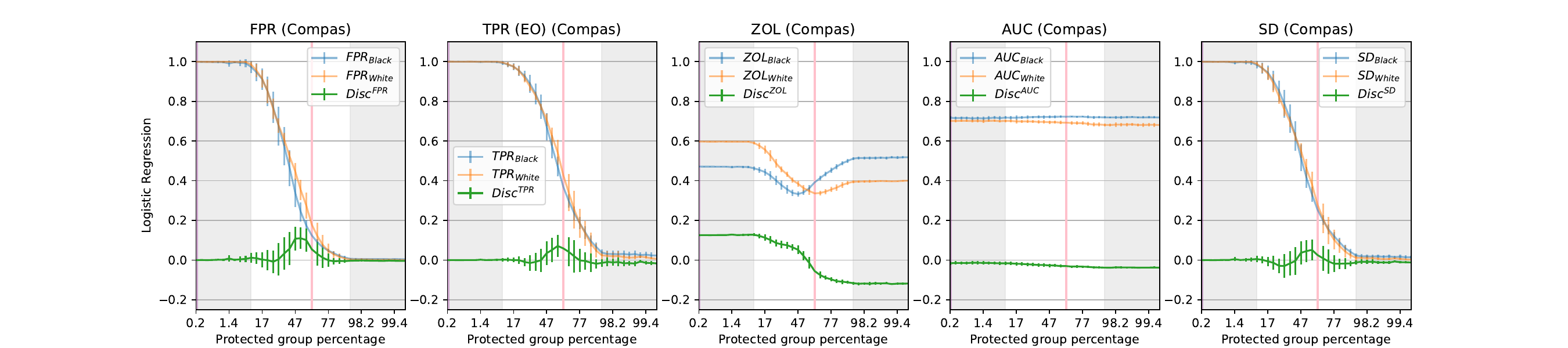}
    \includegraphics[scale=0.38]{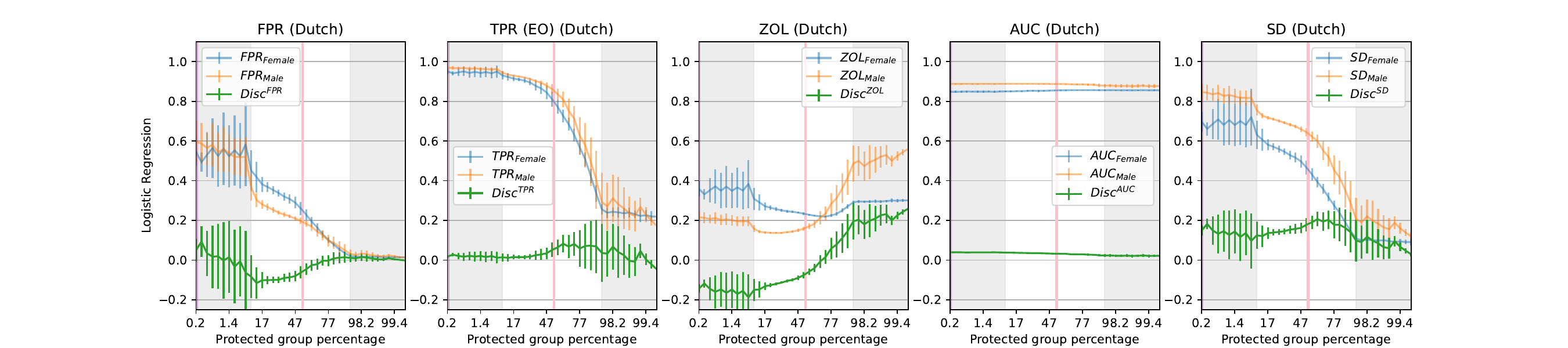}

    \caption{Effect of Bias Mitigation using Reweighing Preprocessing on Discrimination in Logistic Regression for {\em URB} Experiment}
   
\end{figure}




\begin{figure}[H]
    \centering
    \includegraphics[scale=0.38]{Plots/GerryFair/Disc_GerryFairClassifier_0.9_PM_0.1_PF_Adult_LG.pdf}
    \includegraphics[scale=0.38]{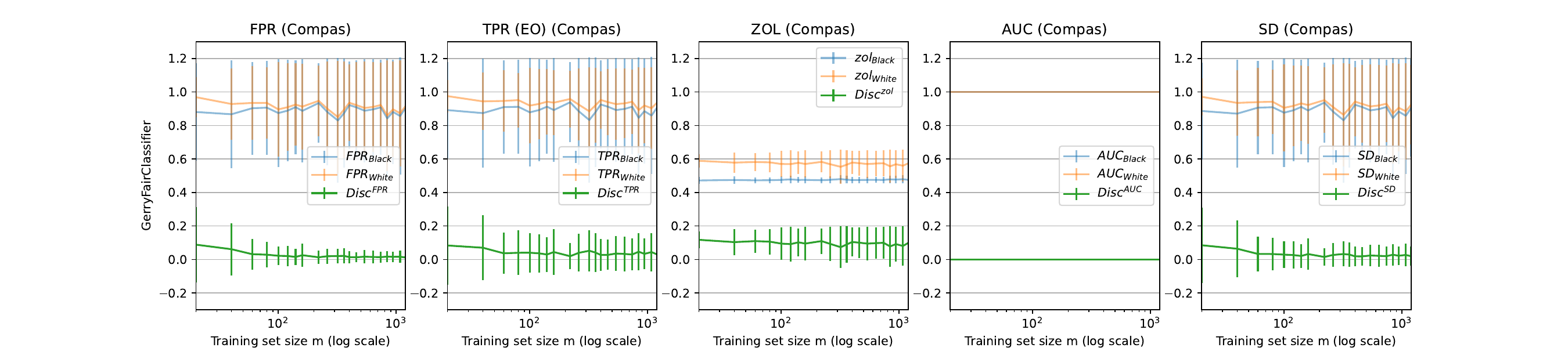}
    \includegraphics[scale=0.38]{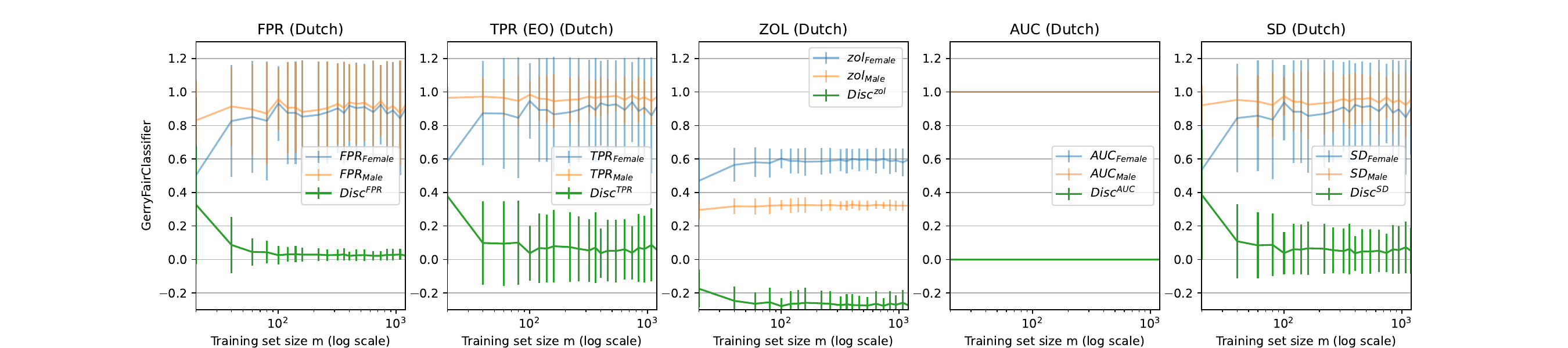}
    \caption{Effect of bias mitigation using GerryFairClassifier in-processing on discrimination for {\em SSB} experiment}
\end{figure}


\begin{figure}[H]
    \centering
    \includegraphics[scale=0.38]{Plots/GerryFair/Disc_GerryFairClassifier_URB_Experience_Adult.pdf}
    \includegraphics[scale=0.38]{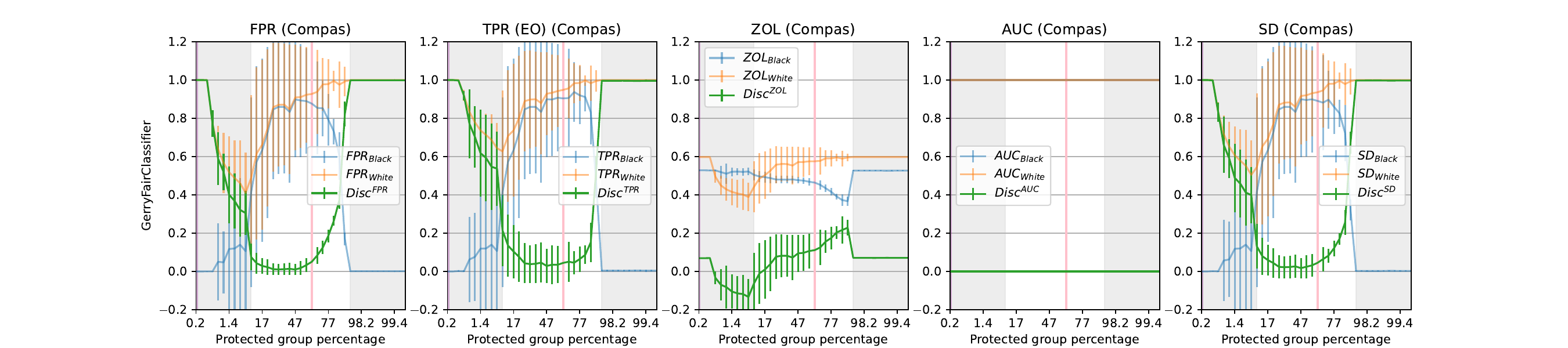}
    \includegraphics[scale=0.38]{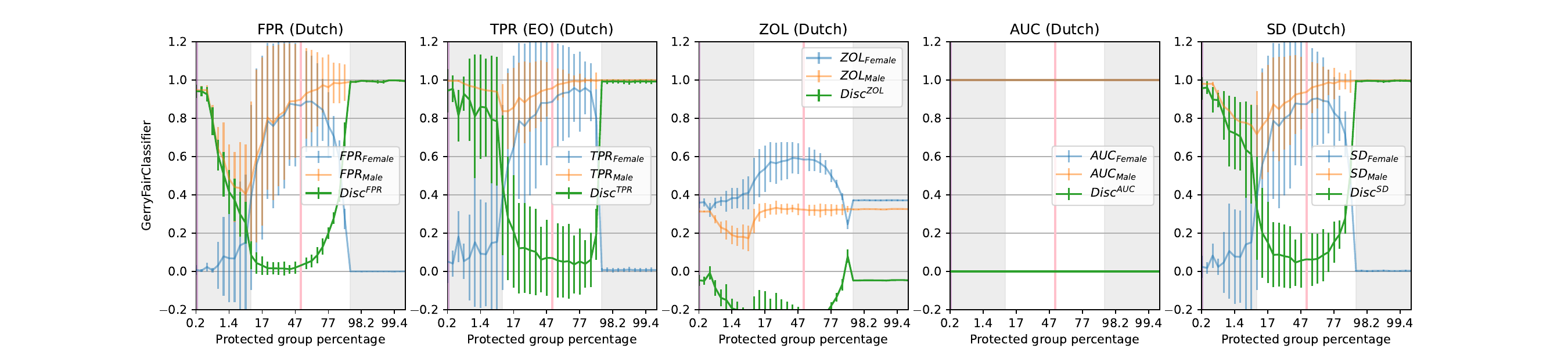}
    \caption{Effect of bias mitigation using GerryFairClassifier in-processing on discrimination for {\em URB} experiment.}
\end{figure}

 \label{sec:a_urb}

\subsection{Additional plots for the effect of data augmentation on discrimination (Section~\ref{sec:exp_threshold})}
\label{sec:a_threshold}

\begin{figure}[H]
    \hspace{-2cm}
    \includegraphics[height=8in, width=7.3in]{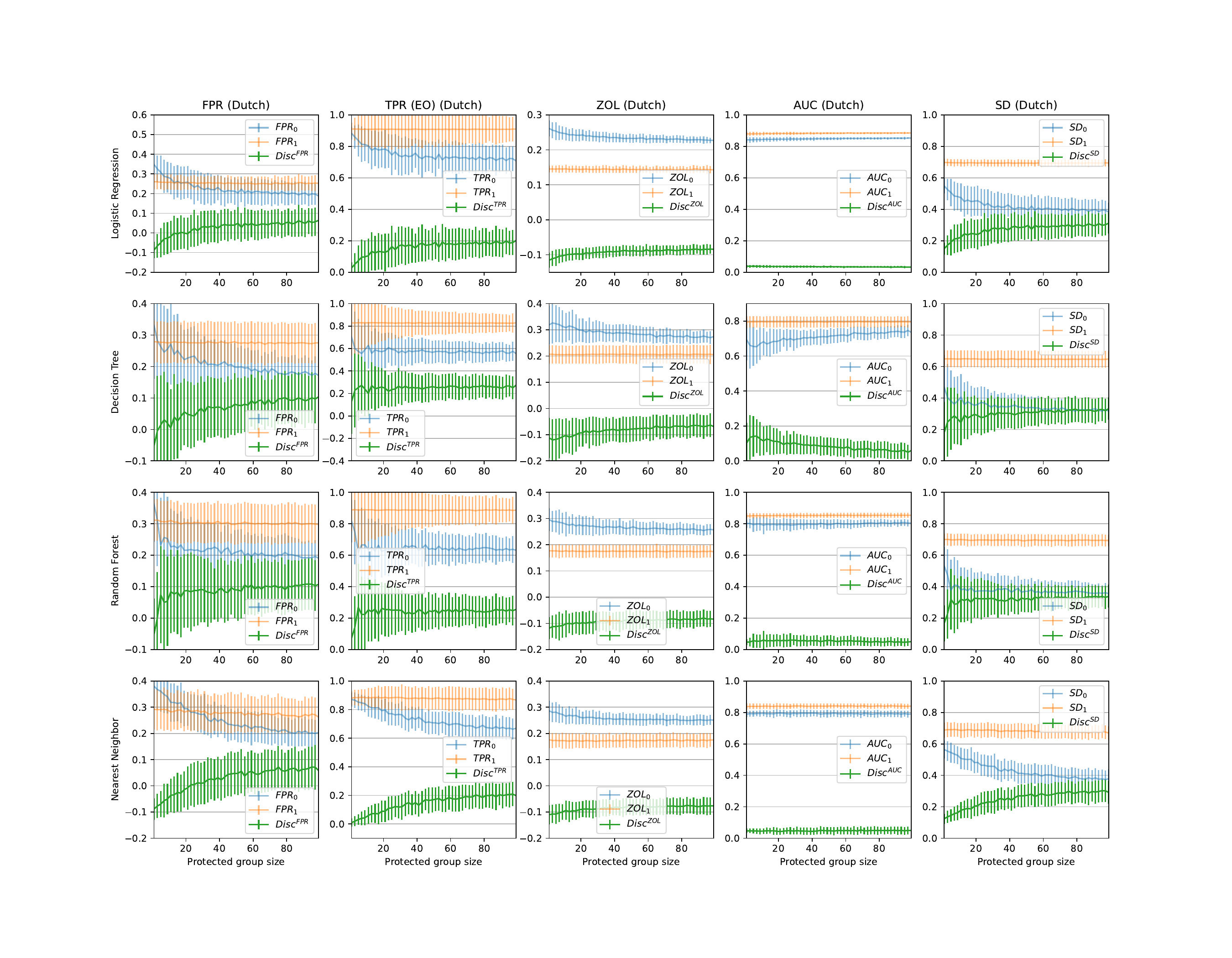}
    \caption{Discrimination values for the Dutch Census dataset while increasing the size of the protected group.}
    \label{fig:DutchThreshold_all}
\end{figure}

\begin{figure}[H]
    \hspace{-2cm}
    \includegraphics[height=7.5in, width=7.1in]{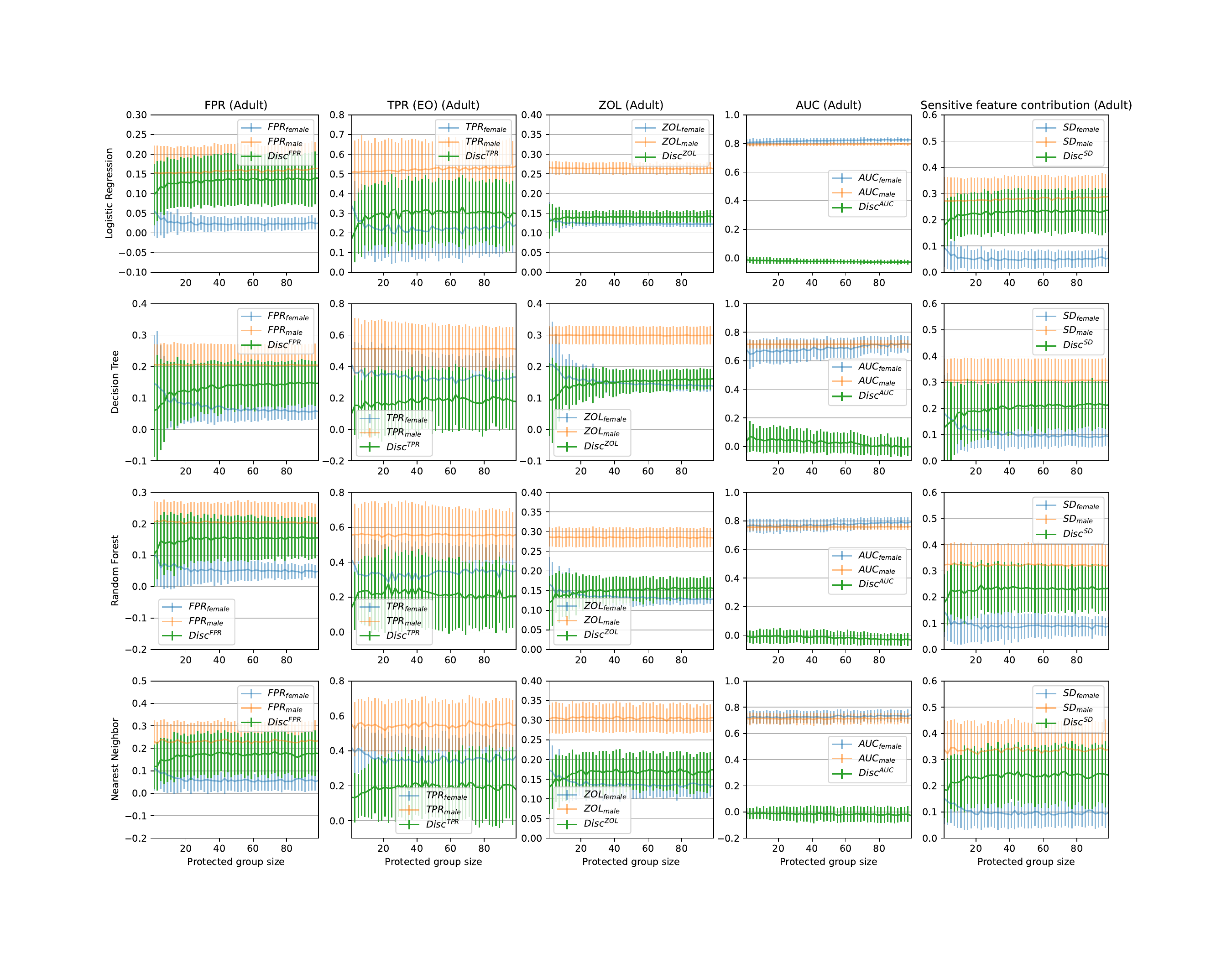}
    \caption{Discrimination value for the Adult dataset while increasing the size of the protected group.}
    \label{fig:adultThreshold_all}
\end{figure}

\begin{figure}[H]
    \centering
    \includegraphics[height=2.3in, width=7in]{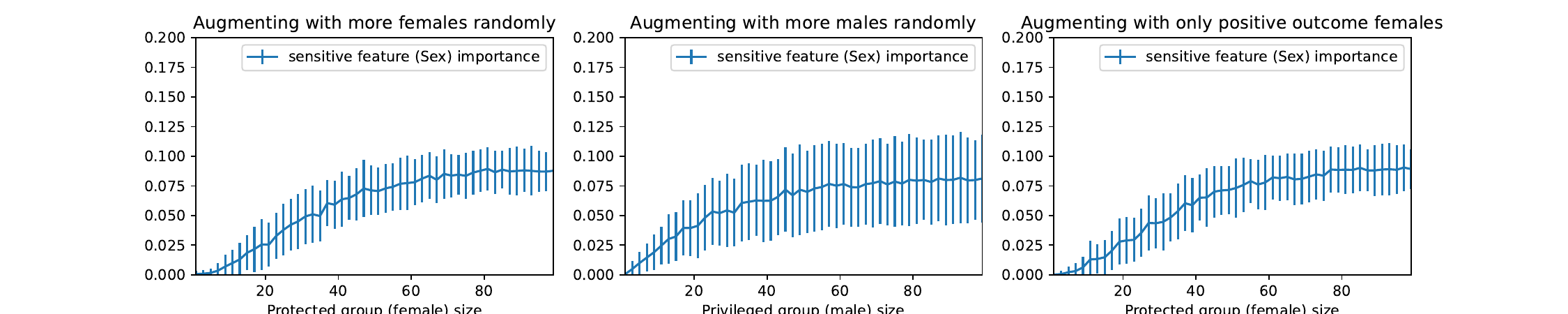}
    \caption{Sensitive feature (Sex) importance observed in the experiments of Section~\ref{sec:exp_threshold}.}
    \label{fig:sensImportance}
\end{figure}

\end{document}